\documentclass{article}

\usepackage{arxiv}

\usepackage[utf8]{inputenc} 
\usepackage[T1]{fontenc}    
\usepackage{hyperref}       
\usepackage{url}            
\usepackage{booktabs}       
\usepackage{amsfonts}       
\usepackage{nicefrac}       
\usepackage{microtype}      
\usepackage{lipsum}

\usepackage[bottom]{footmisc}

\usepackage{setspace} 
\usepackage[utf8]{inputenc} 
\usepackage{amsmath, amsthm, amssymb}	
\usepackage{amssymb}               
\usepackage{graphicx}              
\usepackage{epstopdf}
\usepackage{commath}
\usepackage{lscape}
\usepackage{amsfonts}
\usepackage{icomma} 
\usepackage[small,bf,hang]{caption}

\usepackage{algorithm}
\usepackage[noend]{algpseudocode}

\usepackage{booktabs}
\usepackage{pgfplots}
\usepackage{pgfplotstable}

\author{
	Yanou Ramon\thanks{Corresponding author.} \\
	Dpt. of Engineering Management\\
	University of Antwerp\\
	\And
	David Martens \\
	Dpt. of Engineering Management\\
	University of Antwerp\\
	\And
	Theodoros Evgeniou \\
	INSEAD Paris \\
	\And
	Stiene Praet \\
	Dpt. of Engineering Management\\
	University of Antwerp\\	
}

\begin{document}
	
\title{Metafeatures-based Rule-Extraction for Classifiers on Behavioral and Textual Data}
\maketitle


\begin{abstract}
Machine learning models built on behavioral and textual data can result in highly accurate prediction models, but are often very difficult to interpret. Linear models require investigating thousands of coefficients, while the opaqueness of nonlinear models makes things worse. Rule-extraction techniques have been proposed to combine the desired predictive accuracy of complex ``black-box'' models with global explainability. 
However, rule-extraction in the context of high-dimensional, sparse data, where many features are relevant to the predictions, can be challenging, as replacing the black-box model by many rules leaves the user again with an incomprehensible explanation. To address this problem, we develop and test a rule-extraction methodology based on higher-level, less-sparse ``metafeatures''. 
A key finding of our analysis is that metafeatures-based explanations are better at mimicking the behavior of the black-box prediction model, as measured by the fidelity of explanations.
\end{abstract}

\vspace*{-5mm}
\section{Introduction}
\label{Introduction}
\vspace*{-2mm}

Technological advances have allowed storage and analysis of large amounts of data and have given industry and government the opportunity to gain insights from thousands of digital records collected about individuals each day~\cite{MatzBehavioral}. These ``big behavioral data''---characterized by large volume, variety, velocity and veracity, and defined as data that capture human behavior through the actions and interactions of people~\cite{ShmueliExplainPredict}---have led to predictive modeling applications in areas such as fraud detection~\cite{JellisCustomsFraud}, financial credit scoring~\cite{Martens2007EJOR,DeCnuddeFacebook,TobbackCreditScoring}, marketing~\cite{VerbekeMartensChurn,MatzBehavioral,ChenCloaking} and political science~\cite{PraetPoliticsFB}. Sources of behavioral data include, but are not limited to, transaction records, search query data, web browsing histories, social media profiles, online reviews, and smartphone sensor data (e.g., GPS location data). 
Textual data are also increasingly available and used.
Example text-based applications are automatic identification of spam emails~\cite{spam}, objectionable web content detection~\cite{MartensProvost} and legal document classification~\cite{legalreview}, just to name a few.

Behavioral and textual data are very high-dimensional compared to traditional data, which are primarily structured in a numeric format and are relatively low-dimensional~\cite{JulieExplanations,MatzBehavioral,DeCnuddeBenchmarking}.
Consider the following example to illustrate these characteristics: The prediction of personality traits of users based on the Facebook pages they have ``liked''~\cite{KosinskiFacebook,MatzBehavioral}. A user is represented by a binary feature for each unique Facebook page that exists, with a $1$ if that page was liked by the user and $0$ otherwise, which results in an enormous feature space. However, each user only liked a relatively small number of pages, which results in an extremely sparse data matrix (almost all elements are zero). In the literature, because of their specific nature, behavioral and textual data are often referred to as ``fine-grained''~\cite{MartensProvost,MartensFG,DeCnuddeBenchmarking}. For this reason, in this article we mathematically represent behavioral and textual data as $X_{FG}$ $\subset$ $\mathbb{R}^{n \times m}$, where FG stands for ``fine-grained'', and $n$ and $m$ refer to the number of instances and features respectively. 
These features can be binary (e.g., someone ``liked'' a Facebook page or not) or numerical (e.g., tf-idf vectorization for text documents).

Learning from behavioral and textual data can result in highly accurate prediction models~\cite{JunqueBiggerBetter,DeCnuddeBenchmarking}. A drawback of prediction models trained on these types of data, however, is that they can become very complex. The complexity arises from either the learning technique (e.g., deep learning) or the data, or both. It is essentially impossible to interpret classifications of nonlinear techniques such as Random Forests or deep neural networks.
For linear models or decision trees, the most common approach to understand the model is to examine the estimated coefficients or to inspect the paths from root to leaf nodes. In the context of behavioral and textual data, however, even linear models are not straightforward to interpret because of the large number (thousands to millions) of features each with their corresponding weight~\cite{MartensProvost,JulieExplanations}. Moreover, one may question the comprehensibility of decision trees with thousands of leaf nodes. Alternatively, for linear models, we could inspect only the features with the highest estimated weights. But for sparse data, this means that only a small fraction of the classified instances are actually explained by these features, because of the low coverage of the top-weighted features~\cite{MartensProvost,JulieExplanations}.~\cite{KosinskiFacebook}, for example, explain models that predict personal traits using over $50,000$ Facebook ``likes'' by listing the pages that are most related to extreme frequencies of the target classes. For example, the best predictors for high intelligence include Facebook pages ``\textit{The Colbert Report}'', ``\textit{Science}'' and ``\textit{Curly Fries}''~\cite{KosinskiFacebook}. Because of the extreme sparsity of the data (users liked on average $170$ out of $55,814$ possible pages), these pages are only relevant to a small fraction of users predicted as ``highly intelligent'', which questions the practicality of this approach for better understanding (global) model behavior.

It is important to note that the high-dimensional, sparse nature of behavioral and textual data alone does not necessarily lead to complex prediction models. If many behavioral or textual features are irrelevant for the prediction task, applying dimensionality reduction or feature input selection prior to modeling, or using strong model regularization 
can result in models having high predictive performance, while still being interpretable. However, previous research shows that all of these techniques result in worse predictive performance compared to models that exploit the full set of behavioral or textual features for making predictions~\cite{Joachims1998,JunqueBiggerBetter,ClarkProvost,MartensFG,DeCnuddeBenchmarking,DeCnuddeRamon}. 
By means of a learning curve analysis on a benchmark of $41$ behavioral data sets,~\cite{DeCnuddeBenchmarking} demonstrate that, when mining text or behavior, many features contribute to the predictions. 
Similar results have been found by~\cite{ClarkProvost} and~\cite{JunqueBiggerBetter} for behavioral data, and by~\cite{Joachims1998} for the analysis of textual data. In other words, the dimensionality and sparsity of the data combined with many relevant features drive the ``black-box'' nature of any model trained on behavioral and textual data. We represent a classification model trained on behavioral or textual data $X_{FG}$ as $C_{BB}$, where BB stands for ``black-box''.

Explainability has emerged recently as a key business and regulatory challenge for machine learning adoption. The relevance of global interpretability of classification models is well-argued in the literature~\cite{Andrews1995,Diederich2008,Martens2007EJOR,JunqueALPA}.\footnote{Explanations for model predictions vary in scope: A method either generates global or instance-level explanations~\cite{MartensProvost,RamonSEDC}. We focus on global explanations that give insight in the model's behavior over all possible feature values and for all instances. Instance-level explanations, on the other hand, explain a single model decision. For example, when mining behavior or text, an \textit{Evidence Counterfactual} is an instance-level explanation that shows a minimal set of features such that, when changing their feature values to zero, the predicted class changes~\cite{MartensProvost,RamonSEDC}.} In the process of extracting knowledge from data, the predictive performance of classification models alone is not sufficient as human users need to understand the models to trust, accept and improve them~\cite{VanAsscheBlockeel}. Both the United States and the European Union are currently pushing towards a regulatory framework for trustworthy Articial Intelligence, and global organizations such as the OECD and the G20 aim for a human-centric approach~\cite{EUWhitePaper}. 
In high-stake application domains, explanations are often legally required. 
In the credit scoring domain, for example, legislation such as the Equal Credit Opportunity Act in US Federal Law~\cite{EqualCreditUSLaw} prohibits creditors from discrimination and requires reasons for rejected loan applications. Also in lower-stakes applications, such as (psychologically) targeted advertising~\cite{MatzBehavioral,JulieExplanations} or churn prediction~\cite{VerbekeChurn}, explanations are managerially relevant. Global interpretability allows to verify the knowledge that is encoded in the underlying models
~\cite{Andrews1995,Huysmans2015}. Models trained on big data may learn incorrect trends, overfit the data or perpetuate social biases~\cite{ChenCloaking}. Furthermore, explanations might give users more control of their virtual footprint.
~\cite{MatzPrivacy} argue that insight into what data is being collected and the inferences that can be drawn from it, allow users to make more informed privacy decisions~\cite{MatzPrivacy}. Moreover, global model explainability can help to induce new insights or generate hypotheses~\cite{Andrews1995,ShmueliExplainPredict}.

Rule-extraction has been proposed in the literature to generate global explanations by distilling a comprehensible set of rules (hereafter referred to as ``explanation rules'') from complex classifiers $C_{BB}$. The complexity of the set of rules is largely restricted so that the resulting explanation is understandable to humans. Rule-extraction in the context of big behavioral and textual data can be challenging, and, to our knowledge, has thus far received scant attention. Because of the data characteristics, rule-extraction might fail in their primary task (providing insight in the black-box model) as the complex model needs to be replaced by a set of hundreds or even thousands of explanation rules~\cite{Huysmans2015,Sushil}. 

This article addresses the challenge of using rule-extraction to globally explain classification models on behavioral and textual data. Instead of focusing on rule-extraction techniques themselves\footnote{In this article, we use the decision tree algorithm CART as the rule-extraction algorithm. We leave the comparison of different rule-extraction techniques (Ripper~\cite{Ripper}, C4.5, and so on) as well as more advanced variants like active learning-based rule-extraction \cite{JunqueALPA,Craven1999} to future research. Our main focus is on empirically assessing the value of using metafeatures for extracting explanation rules.}, this article leverages an alternative higher-level feature representation $X_{MF}$ $\subset$ $\mathbb{R}^{n \times k}$, where MF stands for ``metafeatures'' and $k$ represents the number of metafeatures. Metafeatures are expected to improve the fidelity (approximation of the black-box classification model), explanation stability (same explanations for slightly different training sessions - a concept we introduce, which we will be calling just stability) and accuracy (correct predictions of the original instances) of the extracted explanation rules. 
For simplicity, we only focus on classification problems. Our main claim is that metafeatures are more appropriate, in specific ways we discuss, for extracting explanation rules than the original behavioral and textual features that are used to train the model.

This article's main contributions are threefold: (1) We propose a novel methodology for rule-extraction by exploring how higher-level feature representations (metafeatures) can be used to increase global understanding of classification models trained on fine-grained behavioral or textual data; (2) we define a set of quantitative criteria to assess the quality of explanation rules in terms of fidelity, stability, and accuracy; and empirically study the trade-offs between these; and lastly, (3) we perform an in-depth empirical evaluation of the quality of explanations with metafeatures using nine data sets, and benchmark their performance against the explanation rules extracted with the fine-grained features. We aim to answer the following empirical questions: 

\begin{itemize}
	\item How do explanation rules extracted with metafeatures 
	compare against rules extracted with the fine-grained behavioral and textual features across different evaluation criteria (fidelity, stability, accuracy)?
	
	\item How does the 
	fidelity\footnote{As the main goal of rule-extraction is to mimic the behavior of black-box models with a set of rules, we focus on fidelity instead of, say, accuracy, as discussed in Section~\ref{4.4}. Our methodology and analysis can be adapted to also study the accuracy of explanations.} of explanation rules vary over different complexity settings?
	
	\item To what extent do the fidelity and stability of explanation rules extracted with metafeatures depend on a key parameter of our metafeatures-based rule-extraction methodology, that is the parameter $k$ that represents the number of metafeatures?
\end{itemize}


\vspace*{-5mm}
\section{Related work}
\label{RelatedWork_RuleExtraction}

\subsection{Rule-extraction}
\label{RE}
In the Explainable Artificial Intelligence (XAI) literature, rule-extraction falls within the class of post-hoc explanation methods that use ``surrogate models'' to gain understanding of the learned relationships captured by the trained model~\cite{Martens2007EJOR,MurdochInterpretableMachineLearning}. The idea of surrogate modeling is to train a comprehensible surrogate model (the \textit{white-box} $C_{WB}$) to mimic the predictions of a more complex, underlying \textit{black-box} model\footnote{We will interchangeably refer to this model as the black-box model or the underlying model that we want to interpret globally.} $C_{BB}$~\cite{Diederich2008}. We define a black-box model as a complex model from which it is not straightforward for a human interpreter to understand how predictions are made. In this article, we consider \textit{any} classification model (linear, rule-based or nonlinear) utilising a large number of features as a black-box (because of the specic nature of the data described in the Introduction, a large number of features are typically used in the final models); which is different from previous research that only considers highly-nonlinear models as black-boxes~\cite{Andrews1995,Martens2007EJOR,Diederich2008,Martens2009IEEE}.

In the machine learning literature, small decision trees and rule-based models with few rules have been argued to yield the most comprehensible classification models~\cite{VanAsscheBlockeel,Freitas}, making them good candidates to use as white-box models $C_{WB}$ to extract a set of explanation rules (known as ``rule-extraction'').\footnote{
	Note that, in the literature, also linear models with a small number of features have been proposed as surrogate models to approximate a prediction model's behavior~\cite{RibeiroLIME}. In this article, we focus on rule-based models as surrogates.} It is important to note that the complexity of the rules needs to be restricted so that the resulting explanation is comprehensible to humans~\cite{Martens2007EJOR,Martens2009IEEE}. 

Rule-extraction can be used for two purposes. First and foremost, one may be interested in knowing the rationale behind decisions made by a classification model $C_{BB}$ and verify whether the results make sense in practice. The goal is to extract comprehensible rules that closely mimic the black-box, that is measured by what is called ``fidelity''. 
Alternatively, the goal can be to improve the ``accuracy'', namely, the generalization performance of a white-box model (e.g., a small decision tree or a concise set of rules) by approximating the black-box
~\cite{Martens2009IEEE,Huysmans2015}. In this article, we discuss most results in terms of fidelity instead of accuracy as our focus is on developing global explanations that “best mimic the black-box” – but all our analyses can also be done using accuracy as the main metric. 

Rule-extraction methods 
use the mapping of the data to the predicted labels, i.e., the input-output mapping defined by the model $C_{BB}$~\cite{Andrews1995,Martens2007EJOR,Huysmans2015}. The idea behind this approach is that the similarity between the black-box and white-box model (the fidelity) can be substantially improved by presenting the labels predicted by the black-box model $\hat{\textbf{y}}$$=$\{$\hat{y}_{i}$\}$_{i=1}^n$ to the white-box model, instead of the ground-truth labels $\textbf{y}$$=$\{$y_{i}$\}$_{i=1}^n$~\cite{Martens2009IEEE,JunqueALPA}.

\subsection{Challenges of rule-extraction for high-dimensional data}
\label{2.2}

The vast majority of the rule-extraction literature has focused on improving the fidelity and scalability of rule-learning algorithms. However, despite some very impressive and promising work~\cite{Andrews1995,Martens2007EJOR,Martens2009IEEE,Diederich2008,JunqueALPA}, the rule-extraction techniques are mostly validated on low-dimensional, dense data, such as the widely-used set of benchmark data from the UCI Machine Learning repository~\cite{UCIRepository}. These data have feature dimensions going up to $50$ features. We identify at least three challenges in regard to rule-extraction to explain classifiers on fine-grained behavioral and textual data:
\begin{enumerate}
	\item \textit{Complexity of the explanation rules.} In the context of high-dimensional data with many relevant features, rule-extraction might fail to provide insight in the black-box model as the black-box model needs to be replaced by a large set of rules~\cite{Martens2007EJOR,Martens2009IEEE,Huysmans2015}.~\cite{Sushil} apply rule-extraction on (real-world) textual data and show that rule learners can closely approximate the underlying model, but at the cost of being very complex (hundreds of rules). 
	A related challenge that stems from rule-based learners not being very adept at handling high dimensionality, is their high-variance profile that can result in overfitting~\cite{MLalgorithms,DeCnuddeBenchmarking}. 
	\item \textit{Computational complexity.} It is not straightforward for every existing rule-learning algorithm to be used for high-dimensional data, because the learning task can become computationally too demanding~\cite{Andrews1995,Sushil}. Some algorithms, such as Ripper~\cite{Ripper}, are not able to computationally deal with problem instances having large-scale feature spaces~\cite{Sushil}.
	\item \textit{Fine-grained feature comprehensibility.}~\cite{Diederich2008} questions the usefulness of rules extracted from models trained on behavioral or textual data. For example, when rules are learned from a model initially trained on a ``bag-of-words'' representation of text documents, the antecedents in a rule include individual words out of context. This can reduce the semantic comprehensibility of the explanations. Likewise, for digital trace data, we can question the comprehensibility of a single action (e.g., a single credit card transaction, a single Facebook ``like'') taken out of its context, that is, the collection of all behaviors of an individual.
\end{enumerate}

Because of the above challenges, it is questionable \textit{whether fine-grained behavioral and textual features are the best representation for extracting global explanation rules to achieve the best explanation quality in terms of fidelity, stability, and accuracy}. This motivates our approach to use a metafeatures representation instead. It is not clear a priori whether such a representation will improve the quality of explanations of models on behavioral and textual data, making this a key empirical question that we study in this article.

\vspace*{-5mm}
\section{Metafeatures}

\subsection{Motivation}

As previously introduced, behavioral and textual data suffer from high dimensionality and sparsity. For this reason, the features individually may exhibit little discriminatory power to explain the black-box model. 
Because of the low coverage that characterizes such sparse features, a single feature is not expected to ``explain'' much of the classifications of the underlying model. The feature will only be active (nonzero) for a small fraction of all data instances, and therefore, the coverage of an explanation rule with a single behavioral or textual feature is likely to be low~\cite{SommerRuleQuality,MartensProvost,ChenMetaFeatures,Sushil}.\footnote{The coverage of a feature is defined as the number of data instances that have a nonzero value for this feature, whereas the coverage of a rule is defined as the number of instances that are classified by this rule. For sparse data, both feature and rule coverage tend to be low.} 

We address the data sparsity by mapping the fine-grained, sparse features $X_{FG}$ $\subset$ $\mathbb{R}^{n \times m}$ onto a higher-level, less-sparse feature representation $X_{MF}$ $\subset$ $\mathbb{R}^{n \times k}$ (to which we refer as ``metafeatures''), where $m$ and $k$ respectively represent the dimensionality of the original features and metafeatures. Existing research has experimented with the idea of using higher-level features other than the actual features used by the model to extract explanations
~\cite{RibeiroLIME,ChenMetaFeatures,KimTCAV,Lee2019}. In the field of image recognition, for example, the input pixels are not straightforward to interpret, hence researchers have proposed to use a patch of similar pixels (super-pixels) for generating explanations of image classifications~\cite{RibeiroLIME,SuperpixelsImageClassification}. 
Another example stems from the field of natural language processing, where~\cite{ChenMetaFeatures} cope with data sparsity by clustering similar features by their frequency in large data sets. All of these approaches have, however, not been used before in the context of rule-extraction for models on big behavioral and textual data.

\subsection{Desired properties}
\label{sec:desired_properties}

We propose the following set of properties for engineering metafeatures:
\begin{enumerate}
	\item \textit{Low dimensionality.} We want the dimensionality $k$ of the metafeatures to be smaller than the dimensionality $m$ of the original feature space: $k$$<<$$m$. A lower feature dimension may lead to more stable explanation rules~\cite{TommiSelfExplainingNeuralNetworks}. Moreover, the computational burden for extracting rules with metafeatures is likely to be much lower compared to rule-extraction with high-dimensional data~\cite{Andrews1995,Sushil}.
	
	\item \textit{High density.} This property relates to the coverage of a metafeature, which we want to be higher compared to the coverage of fine-grained features~\cite{ChenMetaFeatures}. In other words, there should be more instances for which a metafeature is active (nonzero) compared to the fine-grained features. The higher density (lower sparsity) of the metafeatures is expected to increase the fidelity and accuracy of explanation rules resulting from the higher coverage of rules predicting the non-default target class (often the ``class of interest''). 
	
	\item\textit{Faithful to the original feature representation.} This property is in line with prior research suggesting that the representation of the original data instances in terms of metafeatures should preserve relevant information to discriminate between the predicted labels $\hat{\textbf{y}}$~\cite{TommiSelfExplainingNeuralNetworks}. The metafeatures should preserve the predictive information of the original features as the black-box model is trained on the latter. It is important that the extracted rules using metafeatures can reach a high level of discriminatory power in regard to the true predictions being made, because this will result in a better approximation of the underlying model as measured by fidelity. In the experiments, we use the test fidelity of the explanations as a proxy to measure the faithfulness of the metafeatures $X_{MF}$ to the original features $X_{FG}$. In addition, we use the Gini index to measure the predictive information of each metafeature.\footnote{The Gini index is the splitting criterion of the CART decision tree algorithm that we use in this article to extract explanation rules. The explicit formula is provided in Section~\ref{cognitivelysimple}.}
	
	\item \textit{Semantic comprehensibility.} 
	Metafeatures should have a human-comprehensible interpretation~\cite{TommiSelfExplainingNeuralNetworks}. For example, Facebook ``likes'' can be grouped into semantically-meaningful categories (e.g., ``\textit{Machine Learning}'') and GPS location data can be categorized into venue types (e.g., ``\textit{Concert halls}''). This property is subjective in nature and depends on the application domain and the expectations of users who interact with the model (explanations)~\cite{Wood,Campbell,Huysmans2015,HuysmansRepresentations}. We do not explicitly measure the comprehensibility of explanations with (meta)features, as this would require experimentation with people, an important research direction to explore if indeed metafeatures improve the quality of explanations in the other dimensions we study here (fidelity, stability, accuracy). 
	In this article, we make the assumption that the resulting metafeatures are semantically meaningful. In Section \ref{4.5}, we demonstrate how metafeatures generated with a data-driven method (e.g., Non-negative Matrix Factorization) can be interpreted, based on common practices described in the literature~\cite{WangLatentFactors,OCallLatentFactors,pina2016,MatzLatentFactors,DeCnuddeRamon}. Note that the metafeatures that are manually crafted (the ``domain-based'' metafeatures described in Section \ref{section4.2}) are, by design, comprehensible to humans.
\end{enumerate}


\clearpage
\section{Metafeatures-based Explanation Rules}
\label{RelatedWork}

\begin{figure}[ht]
	\centering
	\scalebox{0.60}{\includegraphics{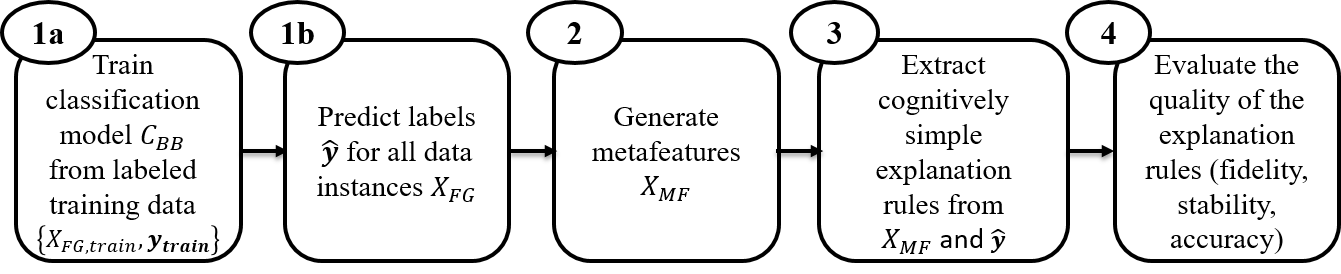}}
	\caption{Proposed rule-extraction methodology using metafeatures.}
	\label{fig:methodology}
\end{figure}

We introduce and validate a methodology to extract explanation rules from a complex model $C_{BB}$ trained on behavioral and textual data $X_{FG}$. The steps of the proposed methodology are summarized in \textbf{Fig.~\ref{fig:methodology}} and discussed below.

\vspace*{-5mm}
\subsection{Model building and predicting labels}
From the behavioral and textual data $X_{FG}$ (having $n$ instances and $m$ features) we train and test the black-box model $C_{BB}$. The model is trained on a subset of $\alpha$$\times$$\beta$$\times$$n$ instances (the training set $X_{FG,train}$ with corresponding labels $\textbf{y}_{\textbf{\textit{train}}}$) and hyperparameters are optimized using a holdout set of $\alpha$$\times$$(1-\beta)$$\times$$n$ instances (the validation set $X_{FG, val}$ with labels $\textbf{y}_{\textbf{\textit{val}}}$). Finally, the generalization performance of the black-box model is evaluated on an unseen part of the data (the test set $X_{FG, test}$ with labels $\textbf{y}_{\textbf{\textit{test}}}$) that contains $(1-\alpha)$$\times$$n$ instances.\footnote{For classification models that do not require hyperparameter tuning, $\beta$ equals $1$. In the experiments in Section $6$, we set $\alpha$ and $\beta$ to $0.8$. Moreover, we use five-fold cross-validation (CV) to evaluate the generalization performance of the black-box model $C_{BB}$ and to measure fidelity and accuracy of the explanation rules $C_{WB}$ on the test data.} The trained black-box model is used to make predictions $\hat{\textbf{y}}$ for all instances in the data set (training, validation and test data), which will thereafter be used to train, fine tune and test our white-box model $C_{WB}$ (the explanation rules).

\subsection{Generating metafeatures $X_{MF}$}
\label{section4.2}
We need to specify a feature transformation process to group behavioral and textual features $X_{FG}$ with similar properties together in metafeatures $X_{MF}$~\cite{ChenMetaFeatures}. 
There are various approaches for generating metafeatures from the original features, either by manually crafting them using domain knowledge~\cite{MurdochInterpretableMachineLearning} or by automatically obtaining them by means of data-driven feature engineering techniques, such as (un)supervised dimensionality reduction. In the following, we use DomainMF and DDMF as abbreviations for domain-based metafeatures and data-driven metafeatures respectively.

\paragraph{Domain-based metafeatures.}
One way of generating metafeatures from the original features is to group features together in domain-based categories that are manually crafted by experts~\cite{MurdochInterpretableMachineLearning,TommiSelfExplainingNeuralNetworks}. For example, for the Facebook ``like'' data, individual Facebook pages can be grouped together in predetermined categories, for example, pages related to ``\textit{Machine Learning}''. This human-selected set of metafeatures can then be used to extract simple rules to explain model predictions, which represent the relative importance of these domain-based metafeatures in the prediction model. In this article, we mathematically denote the domain-based metafeatures as $X_{DomainMF}$ $\subset$ $\mathbb{R}^{n \times k}$. 

\paragraph{Data-driven metafeatures.}
Alternatively, metafeatures can be generated via a data-driven approach, such as matrix factorization-based dimensionality reduction.\footnote{Note that there exist many other techniques that can be explored to generate the metafeatures representation, such as supervised dimensionality reduction (e.g., see \cite{DeCnuddeRamon}) or embedding techniques (e.g., word2vec~\cite{MikolovWord2Vec} or Fasttext~\cite{Fasttext}). Although different methods exist, we do not aim to do a comparison among them in this work. Our focus is to first answer whether metafeatures can help for the problem we study. A positive answer would indicate that it is promising to study other metafeatures methods for this problem in the future. Moreover, if other approaches generate better results, that would only further support our findings about the value of metafeatures for explainability for behavioral and textual data, making our results more conservative.} The idea is to increase density by representing the data in a lower dimensional space without too much loss of information. The original data matrix $X_{FG}$ with $n$ unique instances and $m$ unique features is split into two matrices $L_{n \times k}$ and $R_{k \times m}$ such that: $X_{FG}$$\approx$$L$$R$. The $k$ columns of $L$ are the metafeatures, and each instance will have a representation in the new $k$-dimensional space. The matrix $R$, represents the relationship between the new metafeatures and the original features \cite{ClarkProvost}. 

Metafeatures group together related features. The quality of the metafeatures depends on the number of extracted metafeatures $k$: A value of $k$ that is set too high results in many highly-similar categories, whereas a low value of $k$ tends to generate overly-broad metafeatures. 
The intended goal of generating metafeatures in this article is to use them for rule-extraction, and consequently, we optimize the number of $k$ such that the out-of-sample fidelity of the rules is maximal (we use a validation set to fine tune the value of $k$). We consider values of $k$ from $10$ up to $1000$~\cite{ClarkProvost}. Note that we should not be concerned with generating too many metafeatures because we only need to interpret the ones that are part of the final explanation rules (this is demonstrated in Section $4.5$). 

For generating metafeatures based on matrix-factorization-based dimensionality reduction, we first approximate the original training data $X_{FG}$ by two matrices $L$ and $R$ for a given number of metafeatures $k$ (\textbf{step 1} in \textbf{Fig.~\ref{fig:bigr}} in \textbf{Appendix} $\textbf{1}$). Matrix $R$ maps each metafeature to the original fine-grained features. We ensure mutual exclusivity by transforming matrix $R$ into a binary matrix $R_{\scriptscriptstyle binary}$, where 1 represents the maximum element for each column (fine-grained feature) of $R$ and all other elements are 0 (\textbf{step 2} in \textbf{Fig.~\ref{fig:bigr}}). In other words, each feature only belongs to one metafeature. 
Next, we map the original matrix $X_{FG}$ to $X'$ by multiplying $X_{FG}$ with the transposed binary matrix $R_{\scriptscriptstyle binary}^{T}$ (\textbf{step 3} in \textbf{Fig.~\ref{fig:bigr}}). Finally, matrix $X'$ is normalized over the number of active (fine-grained) features per instance (e.g., total number of behaviors or words) to become matrix $X_{\scriptscriptstyle DDMF-k}$ that represents the metafeatures per instance (\textbf{step 4} in \textbf{Fig.~\ref{fig:bigr}}). We found that the normalized matrix $X_{\scriptscriptstyle DDMF-k}$ produced better results (as measured by test fidelity of the explanations) than utilizing the original matrix $X'$ or even a binary matrix derived from $X'$. We apply the binarization of matrix $R$ to make the explanation rules more interpretable and semantically meaningful. 

In this article, we will experiment with two well-established dimensionality reduction methods based on matrix factorization: Non-negative Matrix Factorization (NMF) and Singular Value Decomposition (SVD). 
NMF is applied in multiple domains to decompose a non-negative matrix
into two non-negative matrices \cite{Lee2001NMF}. In most real-life applications, negative components or subtractive combinations in the representation are physically meaningless. Incorporating the non-negativity constraint thus facilitates the interpretation of the extracted metafeatures in terms of the original data \cite{Wang2012NMF,MatzLatentFactors,ClarkProvost}. 
SVD is a popular technique for matrix factorization across a wide variety of
domains such as text classification \cite{parry2001use} and image recognition \cite{turk1991}. SVD is computed by optimizing a convex objective function and the solution is equivalent to the eigenvectors of the data matrix. 
We will implement these dimensionality reduction techniques using Python's \textit{Scikit-learn} package \cite{scikit2011}. 

An important assumption we make is that the resulting data-driven metafeatures are semantically meaningful. While the obtained metafeatures are not always guaranteed to be interpretable, especially NMF has been shown to provide interpretable results for fine-grained data applications \cite{pina2016,Lee1999NMF}, as compared to other techniques like SVD. Usually, metafeatures are interpreted by looking at the top-weighted features \cite{WangLatentFactors,OCallLatentFactors,pina2016,MatzLatentFactors,DeCnuddeRamon}. It is important to note that only the metafeatures that are part of the final explanation rules need to be interpreted.\footnote{We will demonstrate in Section $4.5$ how to interpret metafeatures that are part of the explanation by looking at the top-weighted features per metafeature.}  

\subsection{Extracting explanation rules}
\label{cognitivelysimple}

Both rule and decision tree learning algorithms can be used for rule-extraction. 
Since trees can be converted into rules, we also use tree algorithms for rule-extraction~\cite{Martens2007EJOR,HuysmansRepresentations,MartensSCI2008}. 
A full review of these techniques is beyond the scope of this article, but we will shortly describe CART\footnote{CART is readily available from the \textit{Scikit-learn} library in Python.}, as this is the technique used in our experiments.

CART can be used for both classification and regression problems and 
it uses the Gini index as a splitting criterion, which measures the impurity of nodes. The best split is the one that reduces the impurity the most. 
We apply CART to the data where the target variable is changed to the black-box predicted class label $\hat{\textbf{y}}$ instead of the ground-truth labels $\textbf{y}$ (see Section \ref{RE}).

The number of extracted explanation rules can be used as a proxy for human comprehensibility.\footnote{A general assumption in the literature is that linear models with few parameters or rule-based models with few rules are more comprehensible than linear models with many parameters or rule sets with many different rules~\cite{Freitas,DOCEvgeniou,HuysmansRepresentations}.} Restricting the complexity of the rule set is also motivated by research on how people make decisions: based on relatively simple rules to avoid excess cognitive effort~\cite{Gigerenzer,DOCEvgeniou} due to cognitive limitations~\cite{SwellerCognitiveLoadTheory}. In the context of consumer decision-making for example,~\cite{DOCEvgeniou} argue that decision rules should incorporate ``cognitive simplicity'': Rule sets should consist of a limited number of rules, each with a small number of antecedents. 
Finally, it is important to note that the concept of comprehensibility in the context of explanation rules comprises many different aspects, such as the size of the explanation, but also the specific application context and subjective opinion and expectations of the end user, which makes it difficult to measure comprehensbility in a generic way~\cite{HuysmansRepresentations,Campbell,Wood}. In line with cognitive simplicity arguments~\cite{DOCEvgeniou,SwellerCognitiveLoadTheory}, in the experiments in Section $6$, we restrict the complexity of the explanations to at most $32$ rules each consisting of at most five antecedents (this is equivalent to a tree depth of at most five).

\subsection{Evaluating explanation rules}
\label{4.4}

\paragraph{Fidelity.}
First and foremost, the explanation rules are evaluated on how well they approximate the classification behavior of the underlying model. Fidelity measures the ability of the rules to mimic the model's classification behavior from which they are extracted. 
Let \{$\mathbf{x}_{i}$,$y_{i}$\}$_{i=1}^n$  represent the labeled data instances and $\textbf{y}^{\textbf{WB}}$ and $\hat{\textbf{y}}$ respectively the white-box and black-box predicted labels. Fidelity is expressed as the fraction of instances for which the label predicted by the explanation rules (the white-box predicted label) equals the black-box predicted label~\cite{Craven1999,Huysmans2015}: 

\begin{equation}
fidelity^{WB} = \dfrac{| \{\hat{y}_{i}=y_{i}^{WB} | \mathbf{x}_{i}\}_{i=1}^n |}{N}
\end{equation}

While most of our analysis is using fidelity, we can extend the fidelity to ``\textit{f-score fidelity}'' (f-fidel), which may be a more appropriate evaluation metric 
when there is a (predicted) class imbalance. The f-fidel is defined as the harmonic mean between $precision$ and $recall$ (w.r.t. the predicted labels $\hat{\textbf{y}}$ rather than the true labels $\textbf{y}$). 
More precisely, the formula of f-fidel is $\frac{2 \cdot precision\cdot recall}{precision+ recall}$, where the $precision$ of the classifier is the fraction of positively-predicted instances that is correctly classified and the $recall$ refers to the fraction of positive instances that is correctly classified as a positive. 

\paragraph{Stability.}
A second important factor is explanation stability - which we will call stability from here on. Users, businesses, or regulators may have a hard time accepting explanations that are unstable (meaning small changes in the data lead to large change in explanations of the black-box model), even if the explanation has shown to have high fidelity and comprehensibility~\cite{VanAsscheBlockeel}.
~\cite{Turney} distinguishes two types of stability: syntactic and semantic stability. Semantic stability is often measured by estimating the probability that two models learned on different training sets, will give the same prediction to an instance. On the other hand, syntactic stability measures how similar two explanations are (e.g., the overlap of features in two different explanations), and is more specific to a particular explanation representation~\cite{Turney}. We argue that syntactic stability is the most relevant type of stability in the context of explaining classification models. To the best of our knowledge, it remains an open question how to measure syntactic stability for different explanation representations, such as rules and trees. We propose a procedure based on the Jaccard coefficient to measure syntactic stability of explanation rules. More specifically, by measuring the overlap of features that are part of the explanations extracted from slightly different subsets of training data.\footnote{The procedure is based on the work of~\cite{FletcherIslamJaccard} who compare sets of patterns using the Jaccard coefficient.} To compute the stability of explanation rules extracted using different data representations $\forall$$X$$\in$\{$X_{FG}$, $X_{DDMF-k}$, $X_{DomainMF}$\} and the black-box predicted labels $\hat{\textbf{y}}$, we propose the following procedure:
\begin{itemize}
	\item \textbf{Step 1}: Generate $B$ samples \{$X_{trainBS,j}$\}$_{j=1}^B$ from the training data $X_{train}$ using bootstrapping.\footnote{In the experiments, we set the number of bootstrap samples $B$ to $10$.}  
	\item \textbf{Step 2}: Extract explanation rules $C_{WB,j}$ from each bootstrap training sample $X_{trainBS,j}$ (this can be the fine-grained or metafeature representation) and the corresponding labels $\hat{\textbf{y}}_{\textbf{\textit{trainBS,j}}}$ predicted by the black-box model. It is important to note that the data-driven metafeatures $X_{DDMF-k}$ need to be computed again for each bootstrap training sample. Obtain $B$ explanations and keep track of the features that are part of the explanations in $B$ sets of features \{$F_{j}$\}$_{j=1}^B$.
	\item \textbf{Step 3}: Make $\frac{B!}{2!(B-2)!}$ 
	pairwise comparisons of the extracted explanations using the Jaccard coefficient. For two sets of features $F_{v}$ and $F_{w}$ (respectively representing the features in explanations $C_{WB,v}$ and $C_{WB,w}$), the Jaccard coefficient is defined as: $J(F_{v},F_{w}) = |F_{v}\cap F_{w}| / |F_{v} \cup F_{w}|$. The Jaccard coefficient equals $1$ if the sets are equal (the explanations have perfect overlap of features) and $0$ if they are disjoint (the explanations are completely different). For the data-driven metafeatures, two metafeatures are considered to be the same when the Jaccard coefficient computed for two metafeatures, as measured over the top features with the highest weight, exceeds a cut-off value of $c$.\footnote{How many top-features to compare between metafeatures and the cut-off value $c$, 
		are parameter values that need to be set in advance. We consider looking at the top-$20$ features in a metafeature and a cut-off value of $c$$=$$0.5$ as suitable choices based on the literature on interpreting factors obtained from dimensionality reduction~\cite{DeCnuddeRamon}.}
	\item \textbf{Step 4}: Compute the average (pairwise) Jaccard coefficient over $\frac{B!}{2!(B-2)!}$ comparisons.
\end{itemize}

Prior research has shown that explanations that rely on high-dimensional data tend to be less robust compared to methods that operate on higher-level features~\cite{TommiSelfExplainingNeuralNetworks}. For this reason, we expect that the extracted explanation rules with metafeatures will be more stable over different training sessions compared to the rules with the original behavioral and textual features. It is important to note, however, that for the metafeatures generated with a data-driven approach ($X_{DDMF-k}$), the computed stability of the explanations also depends on the metafeature generation method (e.g, NMF). For each bootstrap sample, the data-driven metafeatures are computed again. For the domain-based metafeatures $X_{DomainMF}$ and the original features $X_{FG}$, the features do not have to be computed for each bootstrap sample, making this part of the rule-extraction process relatively more stable. 

\paragraph{Accuracy.}
Rule-extraction has also been used to increase the generalization performance of white-box models $C_{WB}$, as measured by accuracy.~\cite{Martens2009IEEE} show that rules that mimic the behavior of an underlying, better-performing model can become more accurate compared to the rules learned from the original data and the corresponding ground-truth labels $\textbf{y}$. Accuracy is defined as the fraction of correctly classified instances by the explanation rules~\cite{Huysmans2015}: 

\begin{equation}
accuracy^{WB} = \dfrac{| \{y_{i}=y_{i}^{WB} | \mathbf{x}_{i}\}_{i=1}^n |}{N}
\end{equation}

\section{Experimental setup}
\label{ExperimentalSetup}

The experiments in this article explore the performance of explanation rules with metafeatures versus the original features on which the model is trained. We make a distinction between domain-based metafeatures ($X_{DomainMF}$) and metafeatures generated with a data-driven method ($X_{DDMF-k}$). The dimensionality reduction parameter $k$ determines the number of metafeatures. The parameter $k$ is fixed for the domain-based, but a hyperparameter for the data-driven metafeatures. 
We evaluate the performance on a suite of classification tasks using nine behavioral and textual data sets. \textbf{Fig.~\ref{fig:experimental_procedure_fidacc}} in \textbf{Appendix} $\textbf{5}$ summarizes the experimental procedure for evaluating the fidelity, f-fidel and accuracy of explanations using five-fold CV, and the explanation stability using bootstrapping. 

\subsection{Data sets and Models}
\label{Data}
Our experimental data comprise seven behavioral and two textual data sets. The data sets are summarized in \textbf{Table~\ref{table:datasets}}. The \textit{Movielens100} and \textit{Movielens1m}~\cite{movielens} data sets contain movie viewing data from the MovieLens website. We focus on the task of predicting the gender of these users. The \textit{Airline} data\footnote{Crowdflower (https://data.world/crowdflower/airline-twitter-sentiment)} contains Twitter data about American airlines, and the task is to predict (positive) sentiment. The Facebook ``like'' data collected by~\cite{PraetPoliticsFB} (\textit{Facebook}) contains likes from over $6,000$ individuals in Flanders (Belgium) and is used to predict gender. The \textit{Yahoomovies} data\footnote{Yahoo Webscope Program (https://webscope.sandbox.yahoo.com/)} also contains movie viewing behavior, from which we predict gender. The \textit{Tafeng} data~\cite{huangTaFeng} consists of fine-grained supermarket transactions, where we predict the age of customers (younger or older than $30$) from the products they have purchased. The \textit{20news} data~\cite{20news} contains about $20,000$ news posts. For this data, the task is to predict whether a post belongs to the topic ``atheism'', based on the words of the post. Another behavioral data set is the \textit{Libimseti} data~\cite{libimseti}, which contains data about profile ratings from users of the Czech social network Libimseti.cz. The prediction task is, again, the gender of the users. Lastly, the \textit{Flickr} data~\cite{flickr} contains millions of Flickr pictures and the target variable is the popularity of a picture (the number of comments it has).

\begin{table}
	\centering
	\caption{Characteristics of the data sets: data type (Type: behavioral/textual), classification task (Target), number of instances (Instances), number of features (Features), number of domain-based metafeatures (DomainMF), balance of the target $b$ (fraction of instances with a positive class label), and sparsity of the data $\rho$ (fraction of zero feature values in the data $X_{FG}$). The data is sorted by increasing number of features $m$.}
	\label{table:datasets}
	\scalebox{1}{
		\begin{tabular}{cccccccc}
			\textbf{Data set} & \textbf{Type} & \textbf{Target} & \textbf{Instances $n$} &\textbf{Features $m$} & \textbf{DomainMF} & \textbf{$b$} & \textbf{$\rho$}\\
			\noalign{\smallskip}\hline\noalign{\smallskip}
			Movielens100 & B & gender & $943$ & $1,682$ & n.a. & $71.05\%$ & $93.69\%$ \\ 
			Movielens1m & B & gender & $6,040$ & $3,883$ & $18$ & $28.29\%$ & $95.76$\% \\ 
			Airline & T & sentiment & $14,640$ & $5,183$ & n.a. & $16.14\%$ & $99.82\%$\\
			Facebook & B & gender & $6,733$ & $5,357$ & $50$ & $32.42\%$ & $98.19\%$\\ 
			Yahoomovies & B & gender & $7,642$ & $11,915$ & n.a. & $71.13\%$ & $99.76\%$ \\ 
			Tafeng & B & age & $31,640$ & $23,719$ & n.a. & $45.23\%$ & $99.90\%$\\
			20news & T & topic & $18,846$ & $41,356$ & n.a. & $4.24\%$ & $99.87\%$\\
			Libimseti & B & gender & $137,806$ & $166,353$ & n.a. & $44.53\%$ & $99.93\%$\\
			Flickr & B & comments & $100,000$ & $190,991$ & n.a. & $36.91\%$ & $99.99\%$\\
			\noalign{\smallskip}\hline
	\end{tabular}}
\end{table}

All data have a high-dimensional feature space with up to hundreds of thousands of features. \textit{Movielens1m}, \textit{Movielens100} and \textit{Airline} have lower-dimensional feature spaces compared to the other data sets. The large sparsity values $\rho$ for all data indicate that the number of active features is very small compared to the total number of features.

\begin{table}
	\centering
	\caption{Average performance of black-box classification models ($\ell_{2}$-LR and RF) on the test data using five-fold CV.} 
	\label{table:models}
	\scalebox{0.9}{
		\begin{tabular}{cccccc}
			\textbf{Data set} & & \textbf{accuracy (\%)} & \textbf{f-score (\%)} &\textbf{precision (\%)} & \textbf{recall (\%)}\\
			\noalign{\smallskip}\hline\noalign{\smallskip}
			Movielens100 & $\ell_{2}$-LR & $72.75$ & $80.87$ & $80.69$ & $81.08$\\
			& RF & $73.17$ & $81.19$ & $80.89$ & $81.52$\\
			\noalign{\smallskip}\hline\noalign{\smallskip}
			Movielens1m & $\ell_{2}$-LR & $78.79$ & $62.60$ & $62.58$ & $63.08$\\
			& RF & $77.10$ & $59.64$ & $59.56$ & $60.15$\\
			\noalign{\smallskip}\hline\noalign{\smallskip}
			Airline & $\ell_{2}$-LR & $89.28$ & $66.62$ & $68.07$ & $70.42$\\
			& RF & $88.05$ & $62.83$ & $64.16$ & $66.37$\\
			\noalign{\smallskip}\hline\noalign{\smallskip}
			Facebook & $\ell_{2}$-LR & $85.22$ & $77.07$ & $77.53$ & $76.83$\\
			& RF & $84.79$ & $76.42$ & $76.82$ & $76.25$\\
			\noalign{\smallskip}\hline\noalign{\smallskip}
			Yahoomovies & $\ell_{2}$-LR & $76.78$ & $83.51$ & $82.70$ & $84.33$\\
			& RF & $76.54$ & $83.46$ & $83.71$ & $83.24$\\
			\noalign{\smallskip}\hline\noalign{\smallskip}
			Tafeng & $\ell_{2}$-LR & $67.69$ & $64.98$ & $67.59$ & $62.55$\\
			& RF & $62.07$ & $57.95$ & $58.19$ & $57.93$\\
			\noalign{\smallskip}\hline\noalign{\smallskip}
			20news & $\ell_{2}$-LR & $96.59$ & $59.83$ & $59.84$ & $60.03$\\
			& RF & $96.58$ & $59.70$ & $59.69$ & $59.92$\\
			\noalign{\smallskip}\hline\noalign{\smallskip}	
			Libimseti & $\ell_{2}$-LR & $82.71$ & $82.89$ & $85.61$ & $89.05$\\	
			& RF & $79.29$ & $81.78$ & $84.62$ & $89.31$\\
			\noalign{\smallskip}\hline\noalign{\smallskip}
			Flickr & $\ell_{2}$-LR &  $82.28$ & $76.00$ & $76.02$ & $76.15$\\ 
			& RF & $81.17$ & $74.50$ & $74.59$ & $74.61$\\
			\noalign{\smallskip}\hline\noalign{\smallskip}
	\end{tabular}}
\end{table}

We train Logistic Regression models with $l2$-regularization ($\ell_{2}$-LR)\footnote{In the literature, Logistic Regression with $\ell_{2}$-regularization has shown to be the best-performing classification model for behavioral and textual data~\cite{DeCnuddeBenchmarking}.} 
with the \textit{Scikit-learn} library (Python). For training the classification models, we use $80$\% of the data, and the remaining $20$\% of the data is used for testing the models. For fine tuning hyperparameters of the model, we use a validation set ($20$\% of the training data). More specifically, the regularization parameter $C$ 
of the $\ell_{2}$-LR model 
is selected 
based on the accuracy on the validation set. 
For preprocessing the text data, we remove stopwords and lemmatize tokens using the \textit{NLTK} package in \textit{Python}, and then use tf-idf\footnote{Tf-idf is short for term frequency and inverse document frequency.} vectorization~\cite{Joachims1998,MartensProvost}. 

Measuring accuracy in practice requires discrete class label predictions, which we obtain by comparing the predicted probabilities to a threshold value $t$ and assigning instances with a predicted probability that exceeds this threshold a positive predicted label. In practice, the choice of the threshold value $t$ depends on the costs associated with false positives and false negatives. In this article, the exact misclassification cost are unknown, and for this reason we compute the threshold value $t$ such that the fraction of instances that are classified as positive equals the fraction of positives in the training data~\cite{Baesens}. \textbf{Table~\ref{table:models}} indicates the generalization performance of all models for each data set over five folds. In addition to the accuracy, we also report the f-score, precision and recall.

To extract explanation rules with the CART algorithm, we use the \textit{DecisionTree} model of the \textit{Scikit-learn} library. For controlling the complexity of the extracted rules, or equivalently the depth of the tree, we change the value of the $max\_depth$ parameter. We let the depth of the tree vary from $1$ to $5$ such that the explanations are cognitively simple (which we motivated in Section~\ref{cognitivelysimple}). 

We extract explanation rules with the original features $X_{FG}$ (on which the classification models are trained) and the metafeatures $X_{MF}$, and the predicted black-box labels $\hat{\textbf{y}}$. In the experiments, we generate data-driven metafeatures $X_{DDMF-k}$ based on two approaches (see Section $4.2$): NMF and SVD. In the experimental results, we mainly discuss the explanations with DDMF generated via NMF (simply denoted by DDMF), that showed the best (fidelity) results. 
For the \textit{Facebook} and \textit{Movielens1m} data, we also extract explanations with domain-based metafeatures.

\section{Experimental results}
\label{ExperimentalResults}
We compare explanation rules for black-box models extracted with metafeatures against those extracted with fine-grained features, across different classification tasks, data sets and evaluation criteria. As mentioned, our main goal is to better understand how metafeatures affect these different criteria and their trade-offs. 




\vspace*{-5mm}
\textbf{\subsection{Are metafeatures better than the original features for explaining models on behavioral and textual data with cognitively simple explanation rules?}}

\begin{table}
	\centering
	\caption{Evaluation of explanation rules for \textbf{$\mathbf{\ell_{2}}$-LR model} using fine-grained features (FG) and data-driven metafeatures (DDMF) with optimal dimensionality reduction parameter $k$ in parentheses. The best performance values (FG vs DDMF) are indicated in bold. The average fidelity, f-fidel and accuracy on the test data are reported over five-fold CV. The stability is measured over $10$ bootstrap samples. The optimal complexity (tree depth) is shown in the last column. We also report the results for explanations with DDMF generated via SVD (\textit{\textit{DDMF-SVD}}). For \textit{Facebook} and \textit{Movielens1m}, we also report results for the domain-based metafeatures (\textit{DomainMF}). 
	}\label{table:CART_table3}
	\scalebox{0.8}{
		\begin{tabular}{ccccccccc}
			\\
			& & Complexity: & tree depth $\leq$$5$\\
			\\
			\textbf{Data set} & \textbf{Representation} & \textbf{fidelity(\%)} & \textbf{f-fidel(\%)} & \textbf{stability(\%)} &\textbf{accuracy(\%)} &\textbf{optimal}\\
			& & & & & & \textbf{depth}\\
			\noalign{\smallskip}\hline\noalign{\smallskip}
			Movielens100 & FG & $72.43$ & $81.99$ & $\mathbf{5.91}$ & $68.93$ & $3$\\
			& DDMF (100) & $\mathbf{75.29}$ & $\mathbf{84.06}$ & $5.52$ & $\mathbf{72.00}$ & $4$\\ 
			& \textit{DDMF-SVD (10)} & $\textit{72.54}$ & $\textit{82.47}$ & $\textit{65.38}$ & $\textit{70.09}$ & $\textit{5}$\\ 
			\noalign{\smallskip}\hline\noalign{\smallskip}
			Movielens1m & FG & $75.53$ & $34.43$ & $8.53$ & $73.29$ & $5$\\
			& DDMF (10) & $\textbf{78.92}$ & $\textbf{57.78}$ & $\mathbf{75.18}$ & $\textbf{74.24}$ & $4$\\
			& \textit{DDMF-SVD (30)} & $\textit{77.89}$ & $\textit{54.12}$ & $\textit{30.81}$ & $\textit{73.08}$ & $\textit{4}$\\ 
			& \textit{DomainMF} & $\textit{71.47}$ & $\textit{21.06}$ & $\textit{46.01}$ & $\textit{70.29}$ & $\textit{3}$ \\
			\noalign{\smallskip}\hline\noalign{\smallskip}
			Airline & FG & $90.53$ & $64.04$ & $\mathbf{33.13}$ & $87.43$ & $4$\\
			& DDMF (700) & $\mathbf{90.79}$ & $\mathbf{65.53}$ & $17.66$ & $\mathbf{88.03}$ & $5$\\
			& \textit{DDMF-SVD (700)} & $\textit{89.62}$ & $\textit{57.88}$ & $\textit{17.66}$ & $\textit{87.08}$ & $\textit{5}$\\
			\noalign{\smallskip}\hline\noalign{\smallskip}
			Facebook & FG & $75.08$ & $41.36$ & $11.71$ & $74.83$  & $5$\\
			& DDMF (70) & $\textbf{81.73}$ & $\textbf{67.99}$ & $\textbf{18.81}$ & $\textbf{79.65}$ & $5$ \\ 
			& \textit{DDMF-SVD (10)} & $\textit{78.30}$ & $\textit{63.74}$ & $\textit{65.32}$ & $\textit{75.91}$ & $\textit{5}$\\ 
			& \textit{DomainMF} & $\textit{77.66}$ & $\textit{59.38}$ & $\textit{50.44}$ & $\textit{76.04}$ & $\textit{5}$\\
			\noalign{\smallskip}\hline\noalign{\smallskip} 
			Yahoomovies & FG & $77.32$ & $85.59$ & $\mathbf{22.62}$ & $72.85$ & $5$\\
			& DDMF (100) & $\textbf{80.41}$ & $\textbf{86.81}$ & $18.91$ & $\textbf{74.05}$ & $5$\\
			& \textit{DDMF-SVD (50)} & $\textit{77.48}$ & $\textit{85.25}$ & $\textit{25.88}$ & $\textit{72.43}$ & $\textit{5}$\\ 
			\noalign{\smallskip}\hline\noalign{\smallskip}
			Tafeng & FG & $68.72$ & $58.27$ & $20.43$ & $57.43$ & $5$\\
			& DDMF (10) & $\textbf{69.39}$ & $\mathbf{62.15}$ & $\textbf{76.62}$ & $\textbf{58.70}$ & $5$\\
			& \textit{DDMF-SVD (300)} & $\textit{69.19}$ & $\textit{62.07}$ & $\textit{14.28}$ & $\textit{57.75}$ & $\textit{5}$\\ 
			\noalign{\smallskip}\hline\noalign{\smallskip}
			20news & FG & $\mathbf{96.38}$ & $\mathbf{32.68}$ &  $\mathbf{26.48}$ & $\mathbf{96.12}$  & $3$\\
			& DDMF (70) & $96.11$ & $27.67$ & $15.78$ & $95.75$ & $3$\\
			& \textit{DDMF-SVD (700)} & $\textit{95.90}$ & $\textit{20.62}$ & $\textit{4.01}$ & $\textit{95.74}$ & $\textit{4}$\\
			\noalign{\smallskip}\hline\noalign{\smallskip}
			Libimseti & FG & $77.18$ & $76.72$ & $24.34$ & $\textbf{94.38}$ & $5$\\
			& DDMF (10) & $\textbf{94.52}$ & $\textbf{94.11}$ & $\textbf{65.39}$ & $87.94$ & $5$\\
			& \textit{DDMF-SVD (10)} & $\textit{93.12}$ & $\textit{92.59}$ & $\textit{68.03}$ & $\textit{99.63}$ & $\textit{5}$\\ 
			\noalign{\smallskip}\hline\noalign{\smallskip}
			Flickr & FG & $63.23$ & $0.84$ & $38.36$ & $63.20$ & $3$\\
			& DDMF (30) & $\mathbf{83.69}$ & $\mathbf{78.09}$ & $\mathbf{38.69}$ & $\mathbf{79.74}$ & $5$\\ 
			& \textit{DDMF-SVD (30)} & $\textit{83.87}$ & $\textit{78.12}$ & $\textit{32.43}$ & $\textit{78.87}$ & $\textit{5}$\\ 
			\noalign{\smallskip}\hline\noalign{\smallskip}
			& \# wins DDMF vs FG & $8-1$ & $8-1$ & $5-4$ & $7-2$\\
			& Mean difference DDMF vs FG ($\sigma$) & $6.05$ ($7.18$) & $16.47$ ($23.87$) & $8.82$ ($37.65$) & $2.40$ ($5.77$)\\
			\noalign{\smallskip}\hline\noalign{\smallskip}
	\end{tabular}}
\end{table}

\textbf{Table~\ref{table:CART_table3}} shows the performance of explanation rules with FG features and metafeatures for the LR models. 
One of the first key questions related to the performance of the rules is ``what is the \textbf{fidelity}'', 
as we want our explanations to mimic the black-box as closely as possible. Overall, the results indicate that the fidelity is higher for DDMF than for the FG-based rules (on average, across all data sets, $6.05$\%). The rules with DDMF achieve a higher number of wins for both the fidelity and f-fidel ($8$ in contrast to $1$). 
We use a one-tailed Wilcoxon signed-rank test~\cite{demsar} to make a statistical comparison between the fidelity of rules with DDMF vs FG features. The test is performed with a sample size of $9$ data sets. 
We find a test statistic $T$$=$$2$ (which is smaller than the critical value $T_{c}$$=$$3$), hence the difference in fidelity between DDMF and FG is statistically significant at a $1\%$ significance level. The difference in f-fidel is statistically significant at a $5\%$ level (test statistic of $T$$=$$5$ compared to a critical value of $T_{c}$$=$$8$). 

One notable exception is the \textit{20news} data: the FG-based rules outperform the DDMF-based rules, and the fidelity values are very high while the f-fidel results are comparably low. This is likely because of the severe (predicted) 
class imbalance ($b$$=$$4.24\%$ in \textbf{Table~\ref{table:datasets}}) compared to the other data. For this reason, the fidelity criterion might be less suitable for this specific data set. Instead, we could have optimized the depth of the tree and the $k$ of the DDMF on the f-fidel as measured on the validation set.\footnote{However, for simplicity, we only used fidelity for all data sets.} 
In order to better understand what drives some of the differences in the performance of explanations with FG features and DDMF, we conjecture this relates to the information held at and the coverage of the most predictive features\footnote{The features are either the fine-grained behavioral or textual features, or the metafeatures. With ``predictive'' we mean predictive in regard to the predicted labels of the black-box model.}. We look at the Gini impurity reduction (used by the CART algorithm) for different features\footnote{We compute the average Gini impurity of the top-FG features and top-MF over five folds.}, which we plot in \textbf{Fig.~\ref{fig:GiniCoverage_LRa}} and \textbf{\ref{fig:GiniCoverage_LRb}} in \textbf{Appendix} $\textbf{6}$. The results (visually) indicate that the ratio in Gini impurity reduction of the top-ranked metafeatures and the FG features relate to the difference in fidelity between rules using FG and DDMF features. For example, consider again the \textit{Flickr} data, for which the explanation rules with metafeatures achieve a fidelity of $20.46$\% higher compared to the FG rules. From \textbf{Fig.~\ref{fig:GiniCoverage_LRa}(g)} we observe that the top-DDMF holds much more information (larger Gini impurity reduction) than the top-FG feature, which might explain the large fidelity difference between the explanations. Indeed, the correlation coefficient between this ratio (impurity reduction of top-ranked DDMF vs top-ranked FG) and the difference in fidelity between explanations with DDMF vs FG (from \textbf{Table~\ref{table:CART_table3}}) is $0.81$.

Secondly, moving to the \textbf{stability} of the explanations, we observe from \textbf{Table~\ref{table:CART_table3}} that the rules with DDMF are similar in stability compared to the FG features ($5$ wins vs $4$). 
The difference in stability is not statistically significant. 
It is important to note that for the DDMF-based explanations, there are two sources of instability: computing metafeatures from different bootstrap samples, and extracting explanation rules for different bootstrap samples. When we would ``fix" the data-driven metafeatures, and not compute them for different bootstrap samples, the stability of the DDMF-based explanations increases, and is comparable to the DomainMF-based explanations. 
Furthermore, the stability results can be closely tied to the parameter $k$. When the optimal dimensionality $k$ of the DDMF is low (for example \textit{Movielens1m} and \textit{Tafeng}), the same DDMF are likely to appear in the global explanation, resulting in more stable explanations over the bootstrap samples. When the selected value of $k$ is higher (for example \textit{20news} and \textit{Airline}), the stability of the explanations with DDMF decreases.

Thirdly, when we compare the \textbf{accuracy} between the rules with DDMF and FG, we observe that the metafeatures-based explanations result in more accurate predictions in regard to the true labels $\textbf{y}$ ($7$ wins versus $2$). However, using a Wilcoxon signed-rank test, we find that the difference in accuracy is not significant at a $5\%$ level. One data set that stands out is \textit{Libimseti}. For this data set, the fidelity and accuracy for DDMF-based explanations as compared to FG-based explanations is respectively better and worse. Stronger even: the accuracies of the explanations with FG, DDMF and DDMF-SVD ($94.38$\%, $87.94$\% and $99.63$\%) are better compared to the accuracy of the black-box model ($82.71$\% in \textbf{Table~\ref{table:models}}). 
Despite the sparsity of this data, there are features that have a large coverage and that are very predictive in regard to the (predicted) target values. For \textit{Libimseti}, there exists a prediction model that has a small number of features (e.g., tree-based model with a depth of $5$) that is more accurate compared to models on the full set of behavioral features. As a consequence, this seems not to be a problem instance that requires post-hoc explanations using rule-extraction. This example illustrates that one should always carefully verify first that there are black-box models on the \textit{full} behavioral or textual data that, indeed, outperform intrinsically-interpretable models (e.g., small decision trees or linear models with a small number of features). If not, it may not help to use a black-box model and then compute post-hoc explanations from it~\cite{Rudin}. Leaving out the \textit{Libimseti} data and performing the Wilcoxon test on the eight remaining data sets, we find that the differences in fidelity and accuracy between DDMF and FG explanations are statistically significant at a $5$\% level.  


Instead of generating metafeatures using a data-driven method, we can also rely on domain-based metafeatures, crafted by experts. The prominent advantage of this approach is that the resulting metafeatures are (by design) comprehensible. However, they may not always be available. For example, we have such metafeatures for only two of the nine data sets: \textit{Facebook} and \textit{Movielens1m}. When comparing DDMF with domain-based metafeatures for these two data sets, we see again that the fidelity is higher for the DDMF compared to the DomainMF (\textbf{Table~\ref{table:CART_table3}} shows that the rules with domain-based metafeatures achieve, at best, test fidelities of $77.66\%$ for \textit{Facebook} and $71.47\%$ for \textit{Movielens1m}), providing further support for using DDMF when developing global explanations for black-boxes. However, when a straightforward semantic meaning of the metafeatures is key, one might still prefer to use DomainMF if they can also increase the fidelity relative to the explanation with FG features (for example for the \textit{Facebook} data). 


\textbf{\subsection{How does the fidelity of explanation rules vary over different complexity settings?}}

\begin{figure}[h]
	\centering
	\scalebox{0.50}{\includegraphics{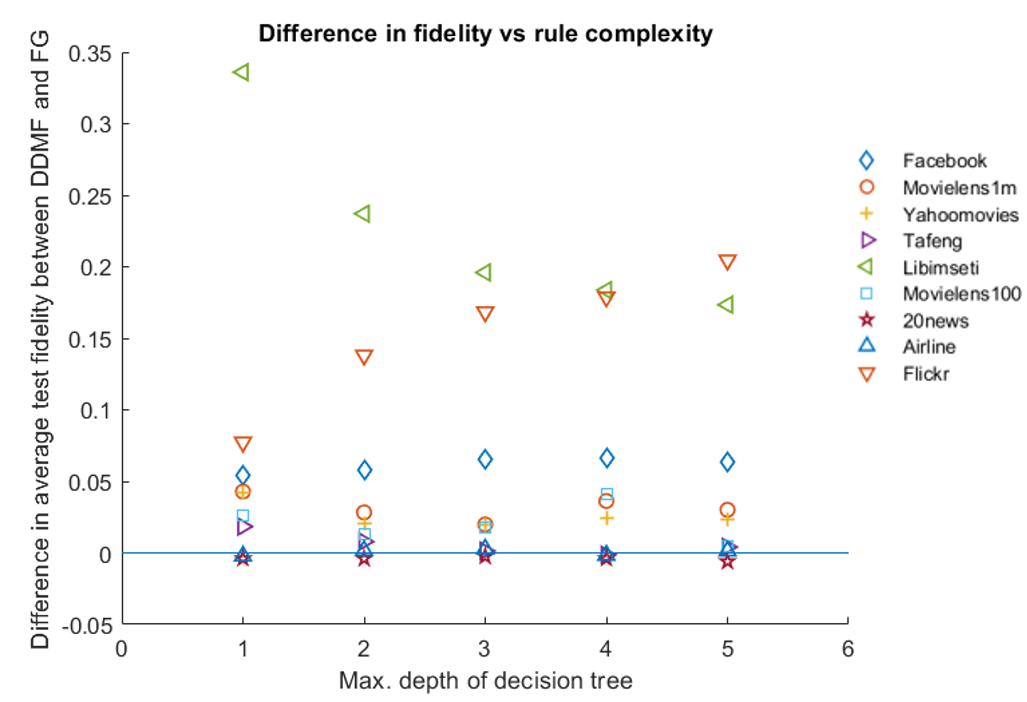}}
	\caption{Difference in average test fidelity of rules with DDMF and FG features in percentage points for varying complexity settings (tree depths from $1$ to $5$).}
	\label{fig:diff_fidel}
\end{figure}

\textbf{Fig.~\ref{fig:diff_fidel}} plots the difference in average test fidelity between rules with DDMF and rules with FG features against the maximum allowed explanation complexity (equivalent to the tree depth). Points above the horizontal line are data sets for which the rules with DDMF perform better. The graph clearly shows that for the majority of data, and over varying complexity settings, the DDMF representation performs better than the FG (differences larger than $0$). Only for the \textit{Tafeng}, \textit{Airline} and \textit{20news} data, the differences are sometimes not positive, indicating that for these complexity settings, the average test fidelity for the rules with FG features is best. In general, from this plot, we can conclude that the findings of \textbf{Table~\ref{table:CART_table3}} hold for varying complexity settings, and that the fidelity is generally higher for explanations with the DDMF representation compared to the FG representation. 

\begin{figure}[ht]
	\centering
	\scalebox{0.55}{\includegraphics{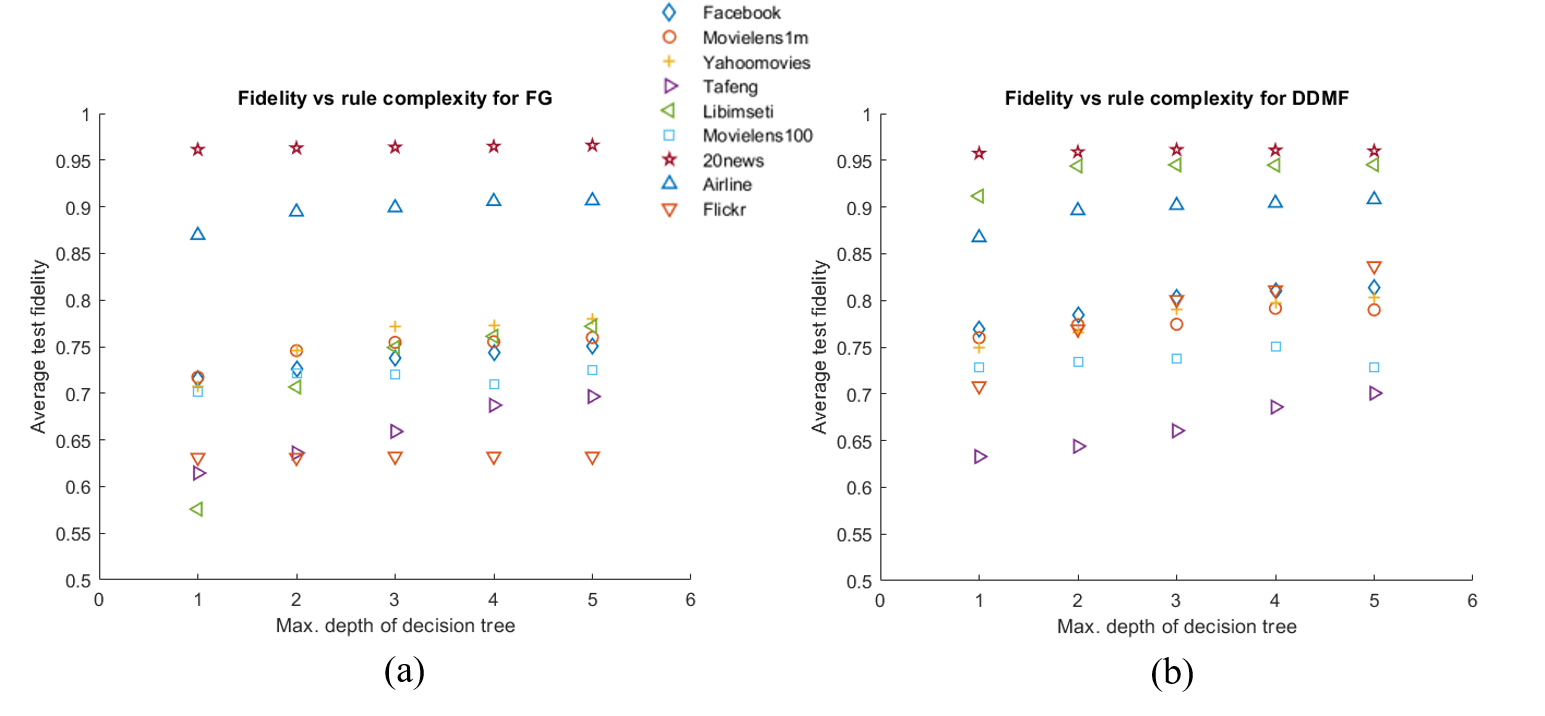}}
	\caption{Average test fidelity of rules with (a) FG features and (b) DDMF for varying complexity settings (tree depths from $1$ to $5$).}
	\label{fig:fidcomplexity}
\end{figure}

\textbf{Fig.~\ref{fig:fidcomplexity}} plots the average test fidelity against the maximal allowed explanation complexity for FG (\textbf{10a}) and DDMF (\textbf{10b}) explanation rules. We observe that, as one would expect, for all data sets, there is generally an increasing fidelity when we increase the depth of the decision tree, or equivalently, the complexity of the explanation rules. 
Interestingly, for some data sets, this fidelity-complexity trade-off is less severe. For example, for the \textit{20news} and \textit{Movielens100} data, the slopes of the curves are relatively flat.
These results also indicate that in some cases, there may not be much to gain by using a relatively ``more complex'' explanation. Therefore, once one is willing to trade-off fidelity for complexity, in some cases, one might as well choose an ``extremely'' simple explanation. 
For the \textit{20news} data, we already pointed at the f-fidel being a more suitable measure than fidelity because of the class imbalance, which might explain the marginal increase in fidelity when increasing complexity.

\textbf{\subsection{How does the number of generated data-driven metafeatures (dimensionality reduction parameter $k$) impact fidelity and stability?}}

A key parameter in our metafeatures-based rule-extraction methodology is the dimensionality reduction parameter $k$. For DomainMF, this $k$ is fixed. For DDMF, we have been selecting the value of $k$ that maximizes the fidelity of the explanation rules on the validation data. As this $k$ may be an important parameter that defines the dimensionality of the space where rule-extraction methods operate (and their performance), we also investigate to what extent the quality - both fidelity and stability - of rules extracted using DDMF depends on this parameter.\footnote{One can do such an analysis for other parameters, too, in general.} Although fidelity can be considered the most important evaluation metric, in practice, one may wish to tune parameters such as $k$ on a desired combination of fidelity, stability and accuracy\footnote{As mentioned earlier, we focus on fidelity - namely how well we can mimic the black-box - instead of accuracy. All analyses can be done for either of the two - or for both – although trade-off decisions become more complex when one uses many criteria.} depending on the context.

\begin{figure}[ht]
	\centering
	\scalebox{0.4}{\includegraphics{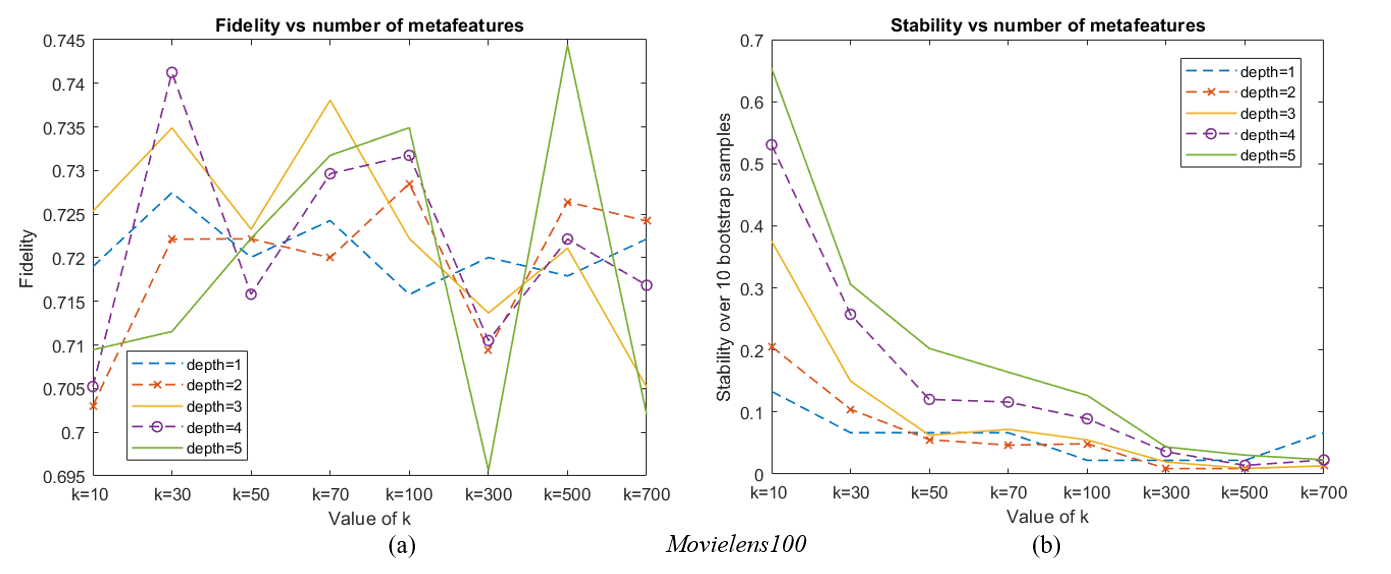}}
	\scalebox{0.4}{\includegraphics{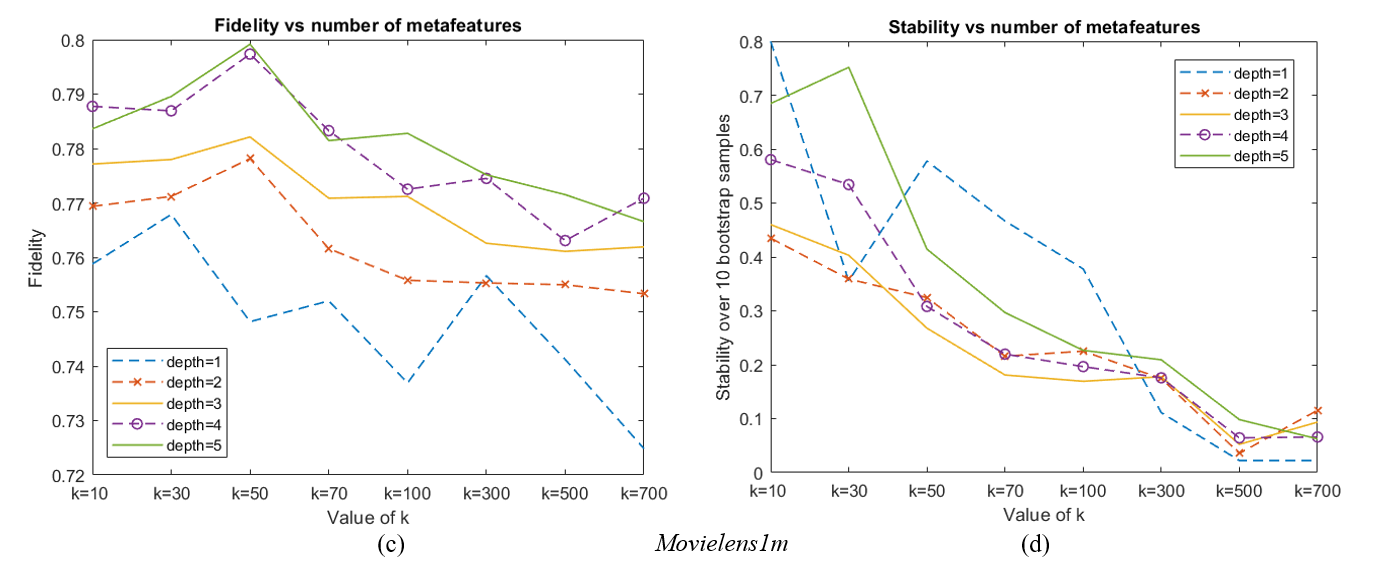}}
	\scalebox{0.4}{\includegraphics{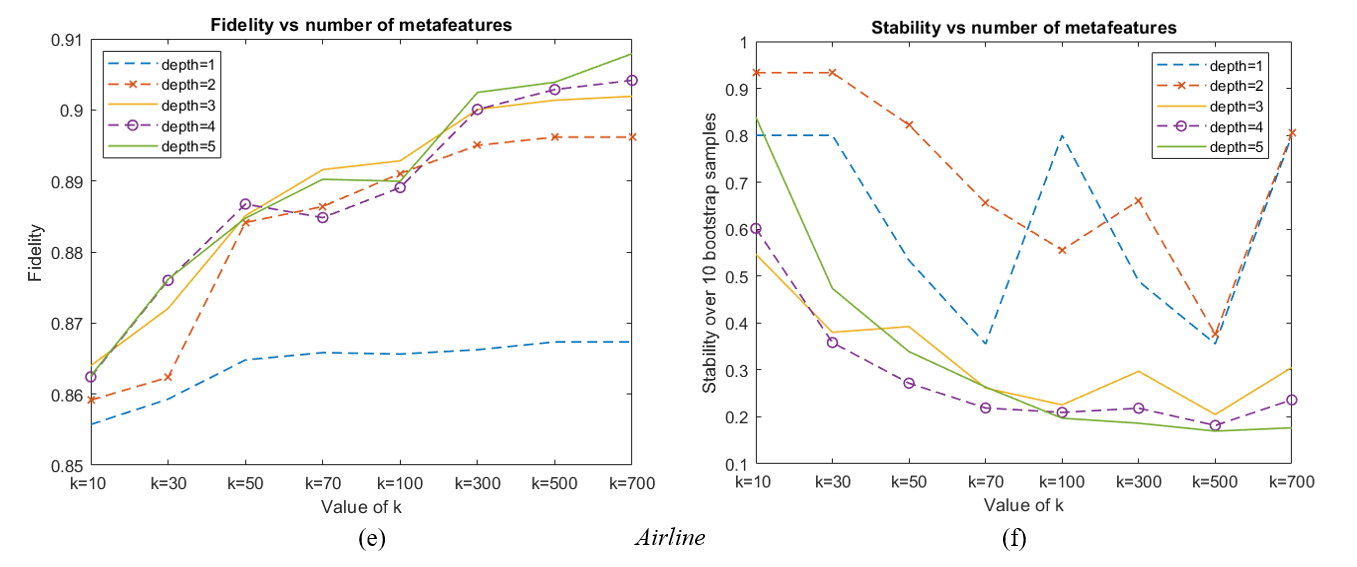}}
	\scalebox{0.4}{\includegraphics{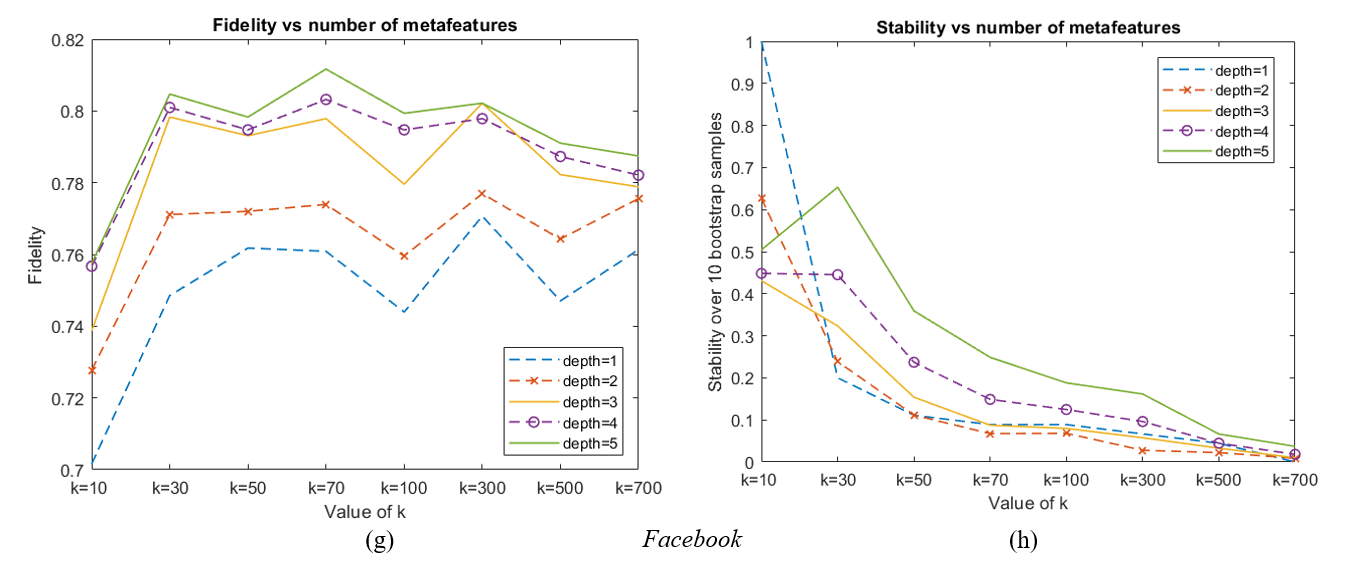}}
	\scalebox{0.4}{\includegraphics{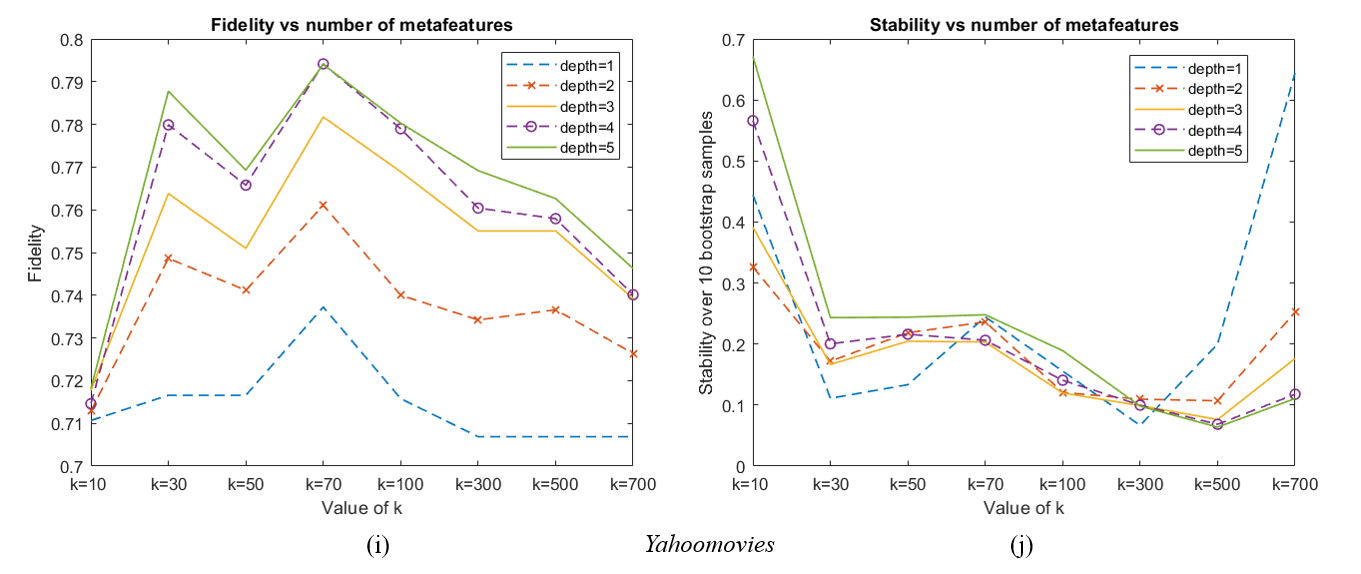}}
	\caption{Average test fidelity and stability for rules with DDMF for varying values of $k$ (number of metafeatures) and complexities for data sets \textit{Movielens100}, \textit{Movielens1m}, \textit{Airline}, \textit{Facebook}, and \textit{Yahoomovies}.}
	\label{fig:numberofmetafeatures}
\end{figure}

\begin{figure}[ht]
	\centering
	\scalebox{0.45}{\includegraphics{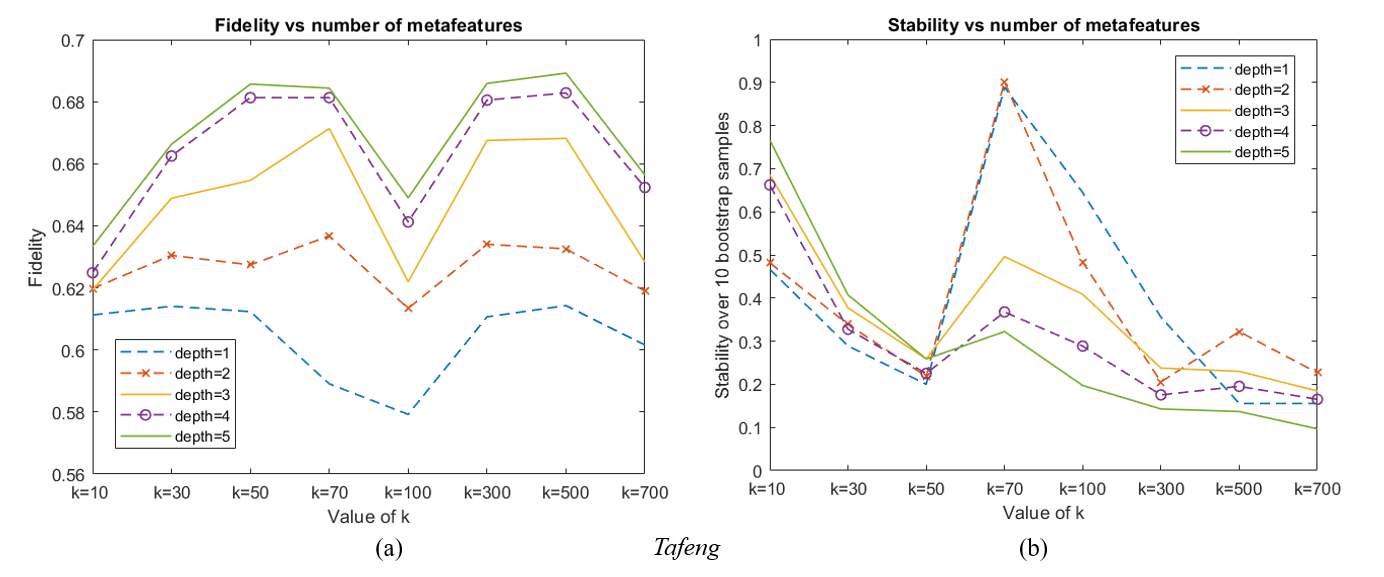}}
	\scalebox{0.45}{\includegraphics{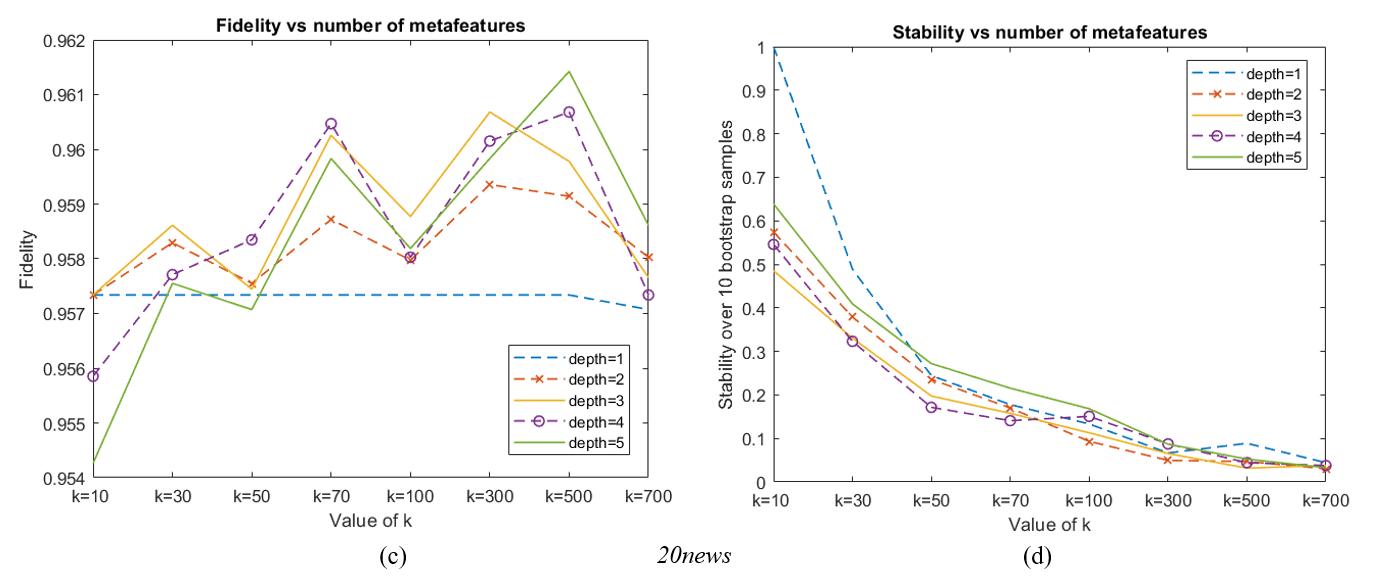}}
	\scalebox{0.45}{\includegraphics{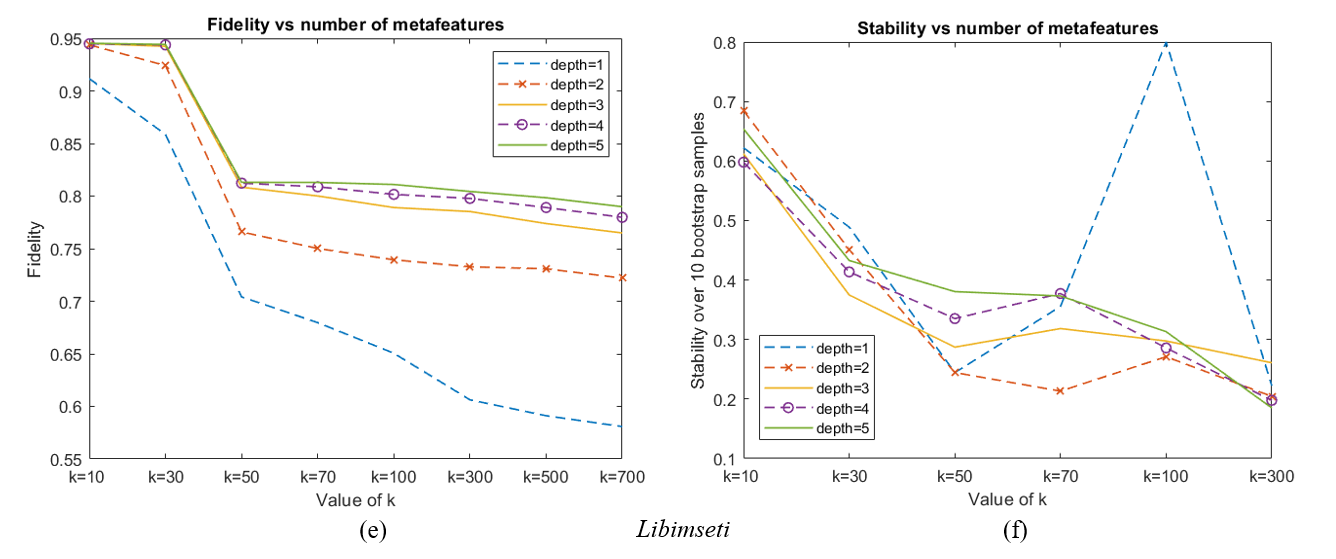}}
	\scalebox{0.45}{\includegraphics{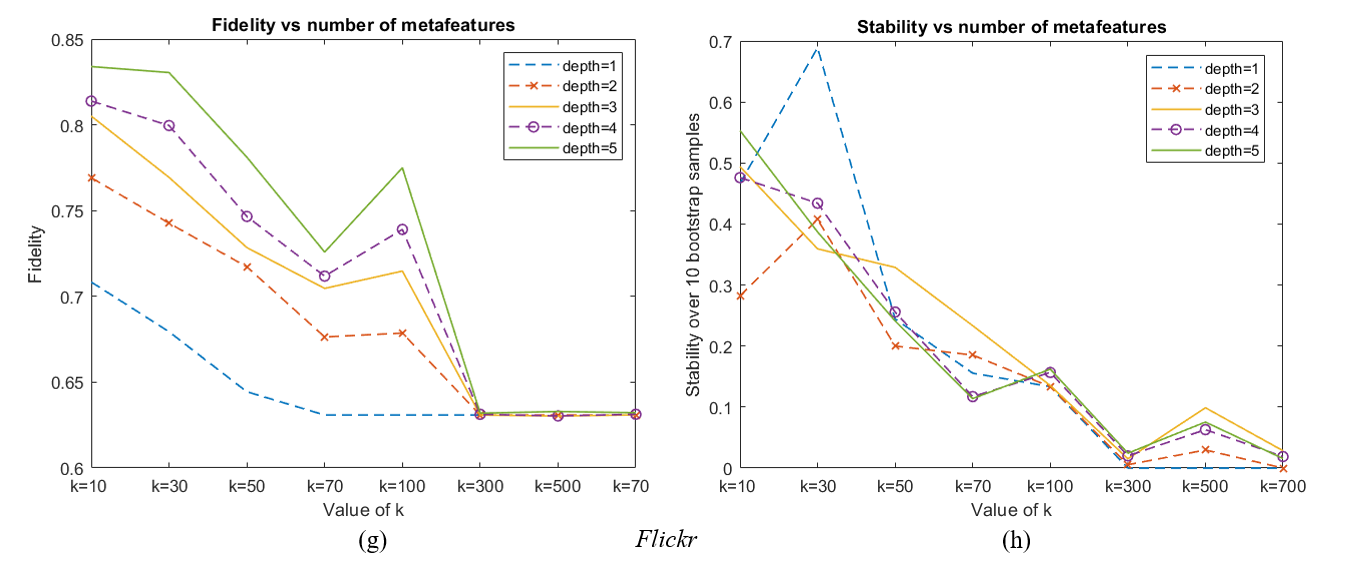}}
	\caption{Average test fidelity and stability for rules with DDMF for varying values of $k$ (number of metafeatures) and complexities for data sets \textit{Tafeng}, \textit{20news}, \textit{Libimseti} and \textit{Flickr}.}
	\label{fig:numberofmetafeatures2}
\end{figure}

\textbf{Fig.~\ref{fig:numberofmetafeatures}} and\textbf{~\ref{fig:numberofmetafeatures2}} show the average fidelity and the stability for varying values of $k$ used and explanation complexities. Firstly, looking at the figures depicted on the left, we observe that, for most data, the fidelity increases with a higher number of metafeatures up until a certain point, after which fidelity decreases again. This turnover point varies per data set, and also depends on the complexity of the explanation (the tree depth). Therefore, an important implication is to select the optimal number of metafeatures on a separate validation set, as we also do. Interestingly, fidelity behaves similarly to how (out-of-sample) accuracy typically does as complexity increases: for both measures there is some sort of ``overfitting'' to the black-box training data in case of including too many metafeatures.

On the other hand, for stability (all figures shown on the right), we observe that, overall, the stability of the extracted rules decreases with a higher number of $k$, especially when allowing for a larger explanation complexity. For example, the stability of rules with DDMF with $k$$=$$10$ is generally larger than the stability with DDMF with $k$$=$$700$. For a lower value of $k$, the dimensionality and sparsity of the metafeatures are lower, making the same metafeature more likely to appear in explanations from different bootstrap samples (as also explained in Section 6.1). 
Interestingly, these figures also show that there is a fidelity-stability trade-off. While fidelity generally increases (at first) with the number of metafeatures and the explanation complexity, stability does not. This may also impact the ``optimal'' number of generated metafeatures $k$, or any parameter selection for any explanation methodology.

\clearpage
\section{Conclusion}
\label{Conclusion}

The fine-grained level of the features that are typically observed in behavioral and textual data sets are of great value for predictive modeling. Feature selection methods or dimensionality reduction techniques to come to a reduced set of ``metafeatures'' have been shown in the literature to lead to lower accuracies~\cite{JunqueBiggerBetter,ClarkProvost,DeCnuddeRamon} for models mining these types of data. On the other hand, we have shown empirically using a number of data sets, and for Logistic Regression 
as black-box model, that these metafeatures are of great value to \textit{explain} the complex prediction models built on the fine-grained features. The results indicate that explanation rules extracted with data-driven metafeatures are better able to mimic the black-box models than those extracted using the fine-grained features. As such, metafeatures help to improve the fidelity: concise rule sets that explain a large(r) percentage of the black-box's predictions (higher fidelity) can be obtained. 

Our empirical results also show important trade-offs between the quality measures of the explanation rules that we considered. For example, more complex explanations (larger rule sets) tend to lead to higher fidelity but lower stability.
An interesting implication of our empirical findings is that one should carefully fine tune any parameter of their explainability method, such as the number of generated metafeatures in our methodology, in order to obtain the desired trade-offs. In our case, increasing the number of generated metafeatures has shown to result in lower stability of the extracted rules, whereas the impact on fidelity is not straightforward and depends on the data set and the complexity setting. 

In this article, we mainly focused on the fidelity of explanation rules in regard to the black-box model. For future research, there are some other important directions to explore for evaluating post-hoc explanations of prediction models: the computational cost to achieve the explanations, the cost of having an explanation rule set with an accuracy that is lower than the black-box model, or the issue of presenting only one rule set as explanation, while other rule sets with similar fidelity and accuracy might exist. Although these aspects are implicitly addressed in our article, a more qualitative study on how these ``costs'' are perceived by users can be another interesting issue for future research. On a methodological level, this study could spur future research on the use of other feature engineering techniques such as embeddings to be used in metafeatures-based explanation rules. One interesting approach is to include the fidelity, stability, accuracy, and complexity measures explicitly when constructing the metafeatures.

Finally, our metafeatures-based explanation approach for high-dimensional, sparse behavioral and textual data has important practical implications for any setting where such data is available and explainability is an important requirement, be it for model acceptance, validation, insight, or improvement. This article could therefore potentially lead to a wider use of valuable behavioral and textual data in domains like marketing and fraud detection among others.

\clearpage
\section{Appendix}

\subsection{Appendix 1: Procedure for generating data-driven metafeatures.}

\begin{figure}[ht]
	\centering
	\scalebox{1}{\includegraphics{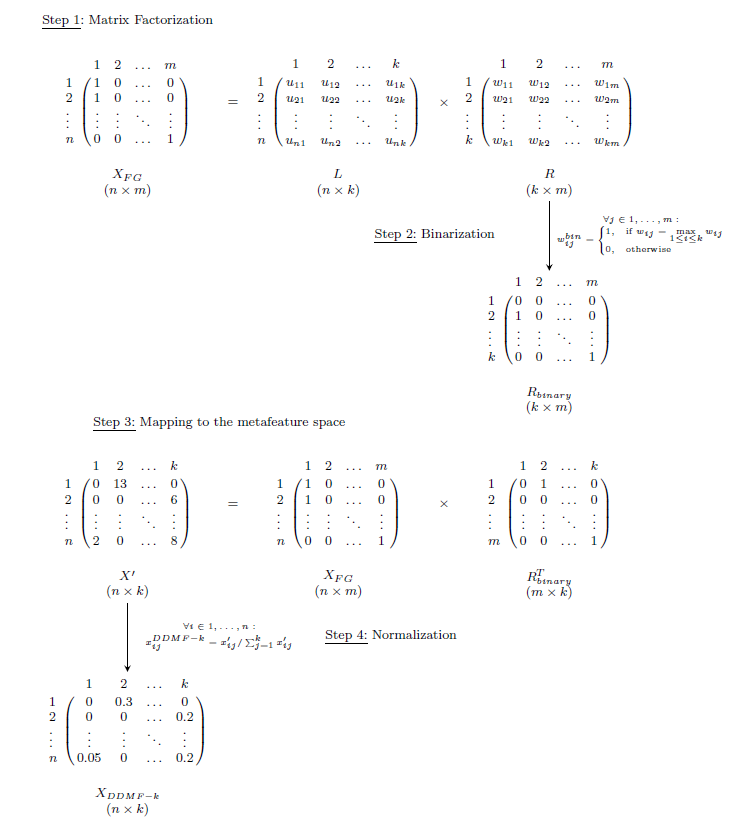}}
	\caption{Procedure for generating DDMF using dimensionality reduction with matrix factorization. Note that $X_{FG}$ can also contain numerical features.}
	\label{fig:bigr}
\end{figure}

\clearpage
\subsection{Appendix 2: Experimental procedure for evaluating fidelity, f-fidel, accuracy, and stability of explanation rules.}
\begin{figure}[ht]
	\centering
	\scalebox{0.8}{\includegraphics{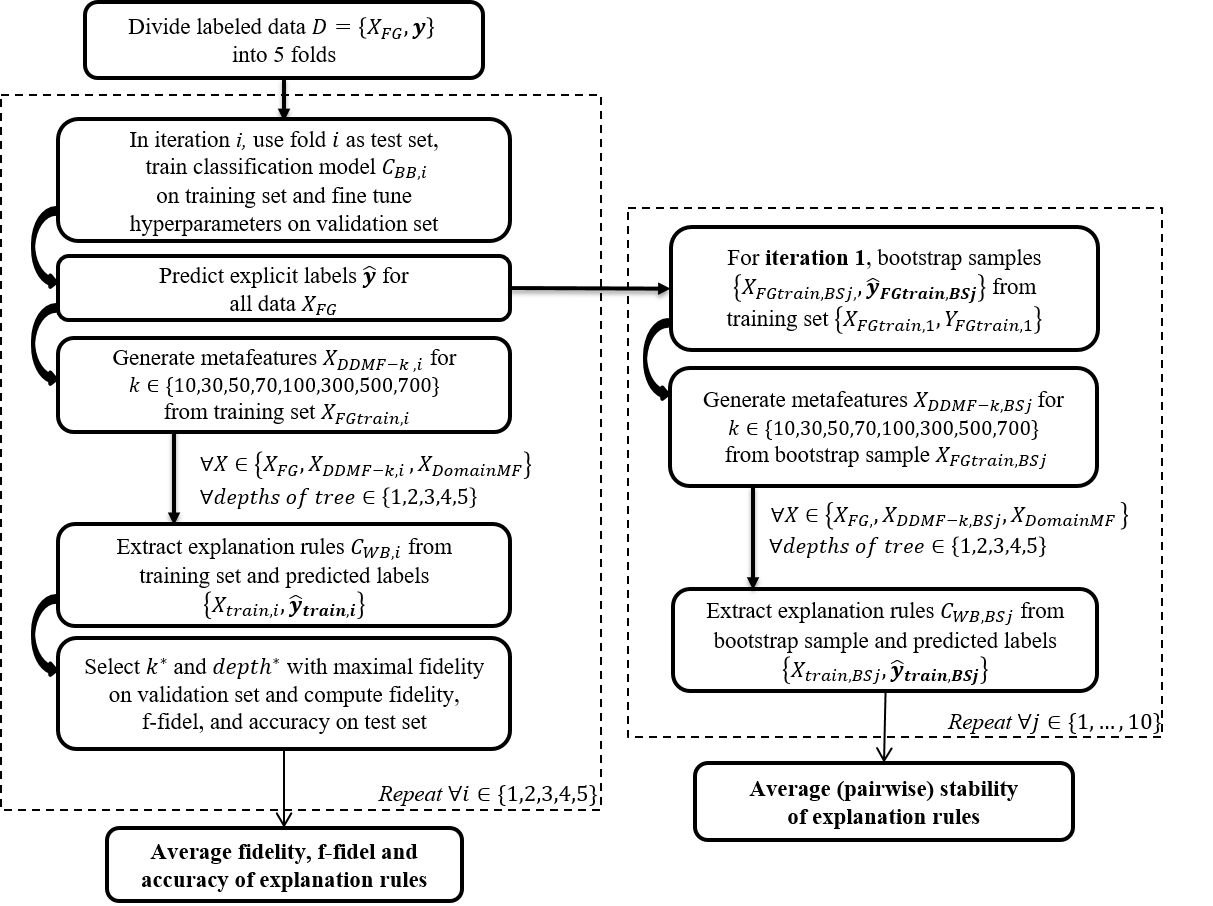}}
	\caption{Experimental procedure of evaluating fidelity, f-fidel and accuracy of explanation rules $C_{WB}$ with five-fold CV, and stability using bootstrap samples, using fine-grained features (FG), data-driven metafeatures (DDMF) and domain-based metafeatures (DomainMF), and varying complexity settings for explanations (this is equivalent to the tree depth).}
	\label{fig:experimental_procedure_fidacc}
\end{figure}

\clearpage
\subsection{Appendix 3: Gini impurity reductions.}

\begin{figure}[ht]
	\centering
	\scalebox{0.5}{\includegraphics{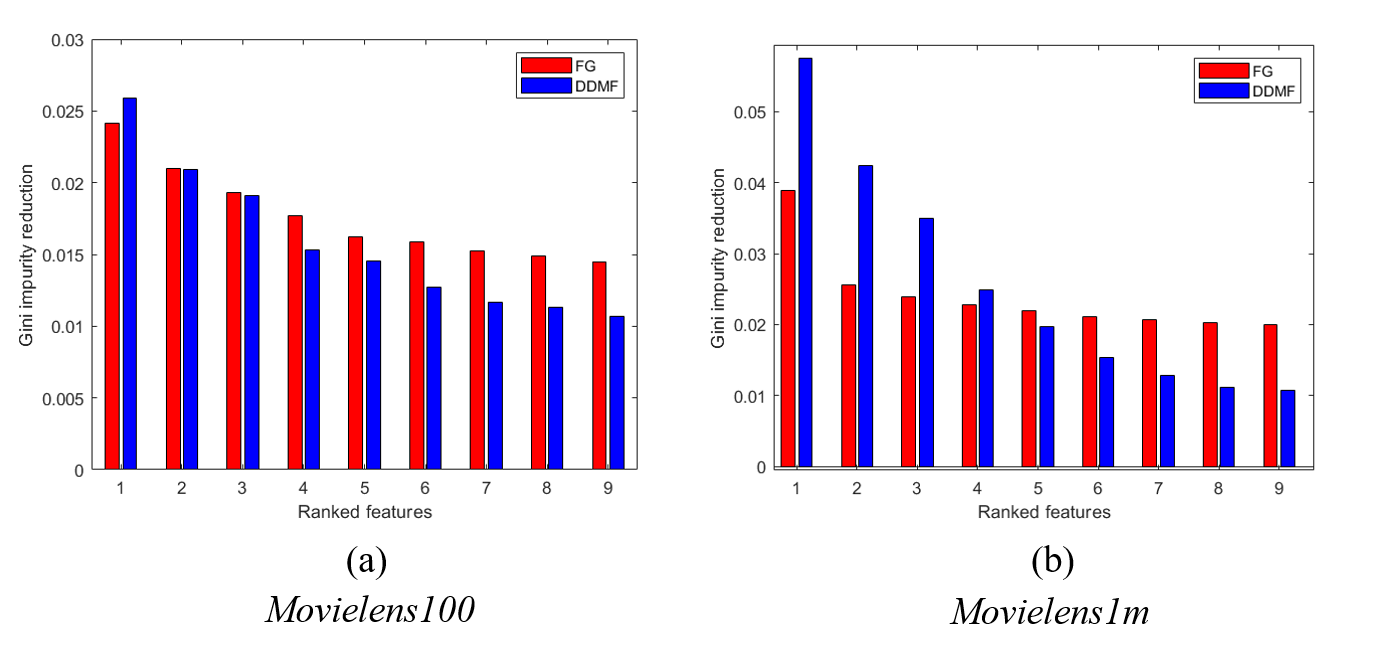}}
	\scalebox{0.5}{\includegraphics{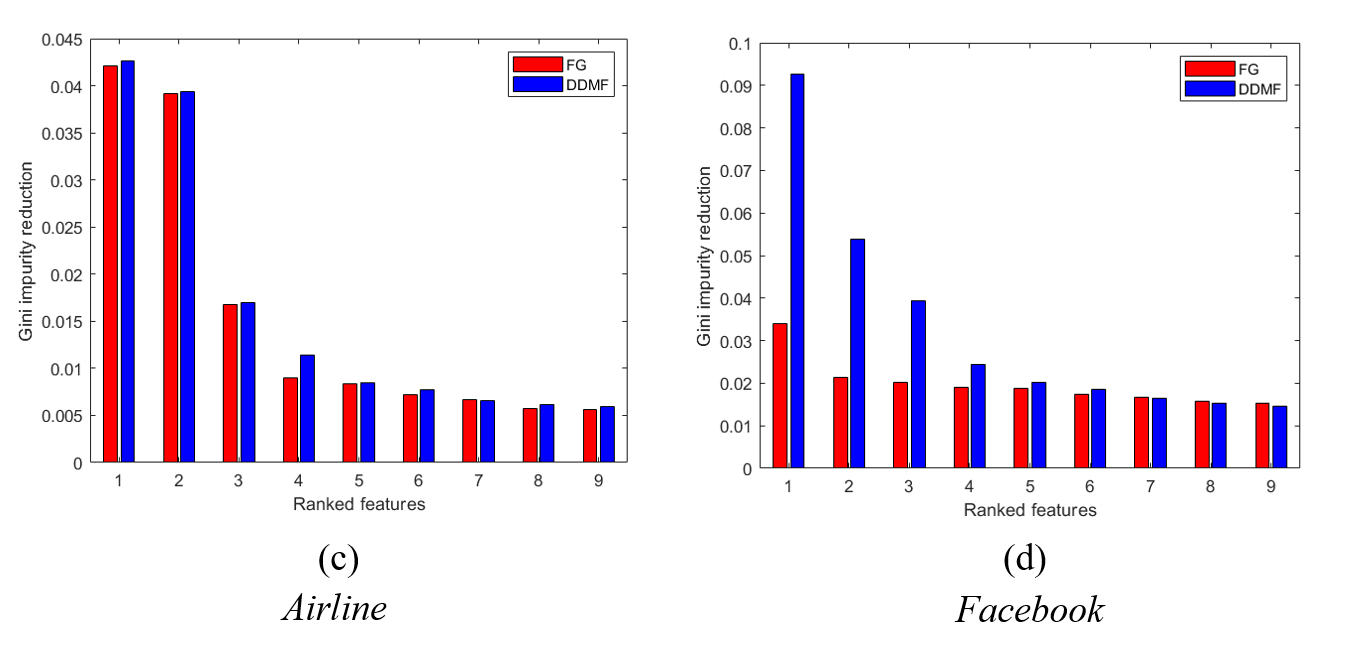}}
	\scalebox{0.5}{\includegraphics{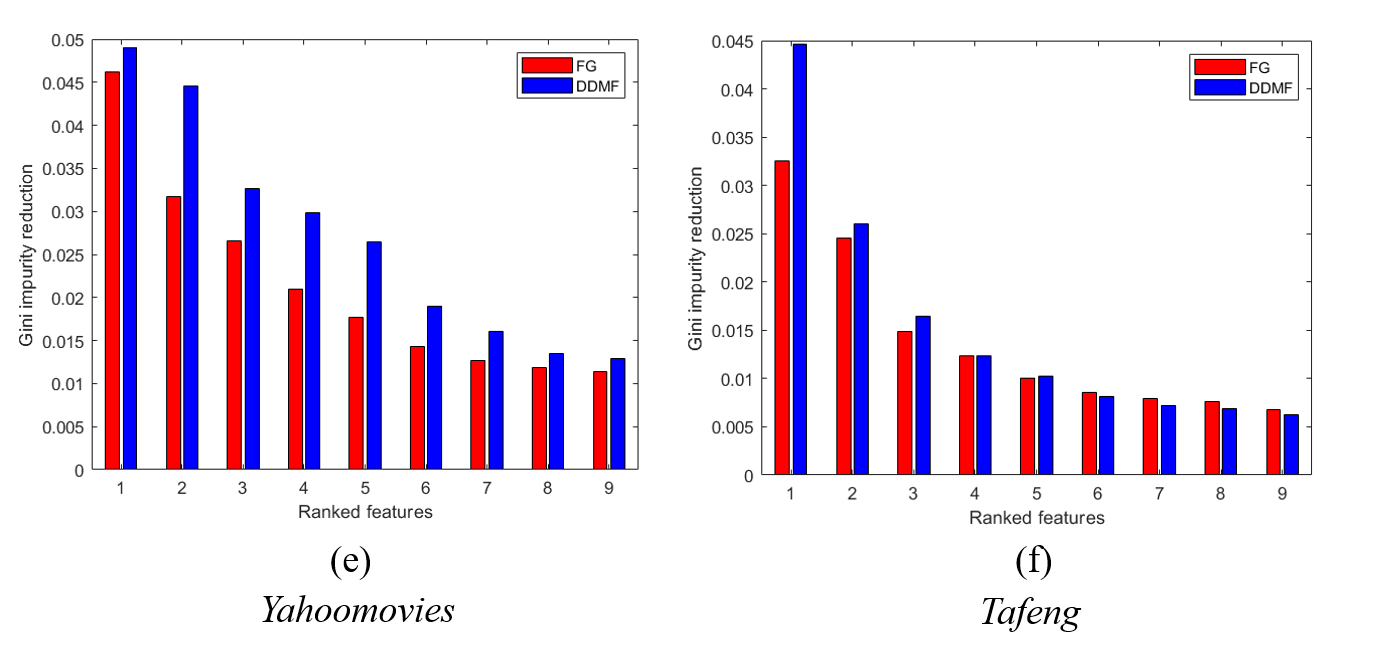}}
	\caption{Top-ranked features with highest Gini impurity reduction for each data representation for the $\ell_{2}$-LR model as the black-box model. The reductions are averaged over five folds.}
	\label{fig:GiniCoverage_LRa}
\end{figure}

\begin{figure}[ht]
	\centering
	\scalebox{0.5}{\includegraphics{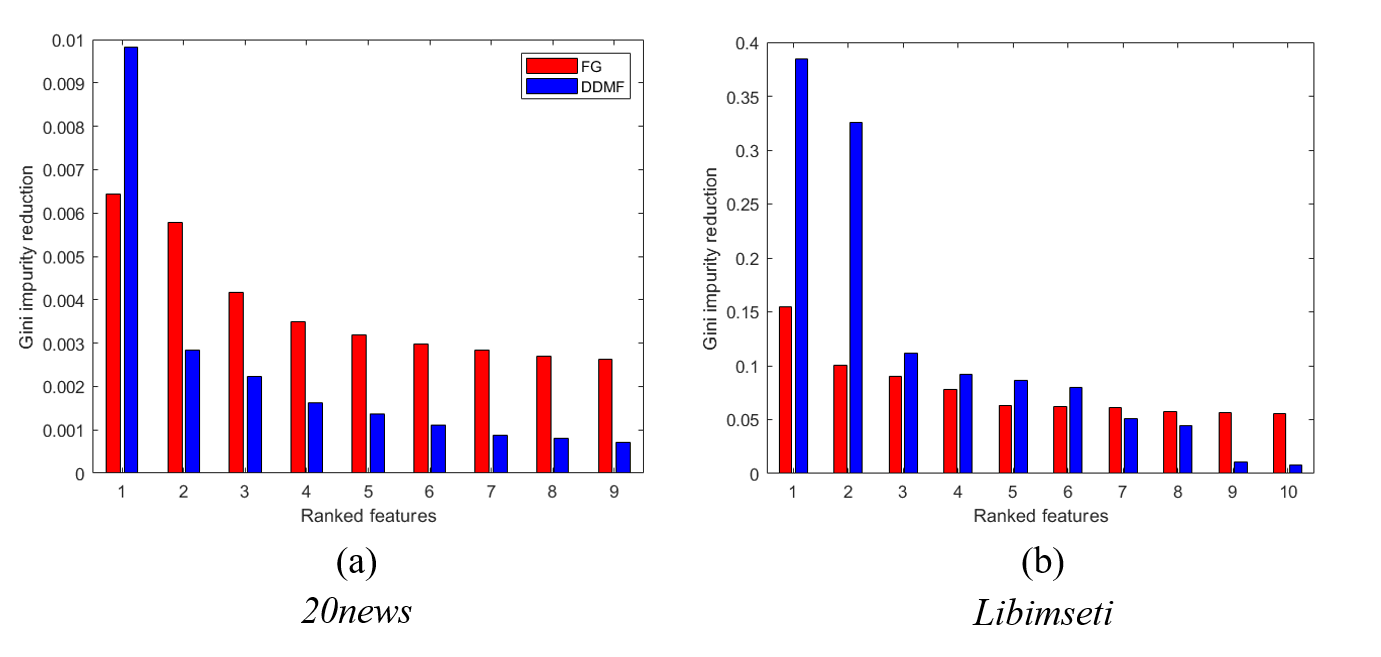}}
	\scalebox{0.5}{\includegraphics{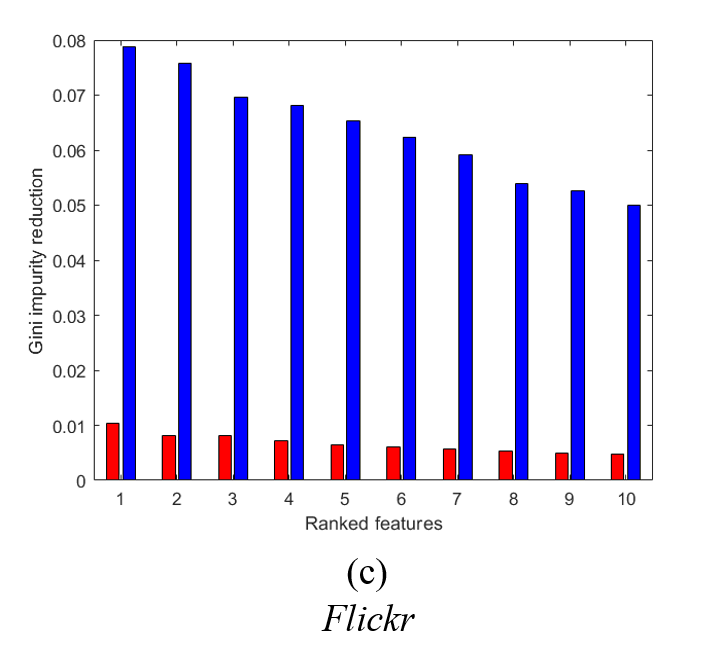}}
	\caption{Top-ranked features with highest Gini impurity reduction for each data representation for the $\ell_{2}$-LR model as the black-box model. The reductions are averaged over five folds.}
	\label{fig:GiniCoverage_LRb}
\end{figure}

\end{document}